\documentclass[letterpaper]{article} 
\usepackage[preprint]{aaai2027}  
\usepackage[hyphens]{url}  
\usepackage{graphicx} 
\usepackage{natbib}  
\usepackage{caption} 
\usepackage{booktabs}
\usepackage{amsmath}
\usepackage{amssymb}
\usepackage{fancybox}
\usepackage{tikz}
\usepackage{pifont}
\usepackage{xcolor}

\newcommand{\rowcolor}[1]{}
\newcommand{\greenstep}[1]{%
  \tikz[baseline=(char.base)]{
    \node[circle,fill=green!50!black,
      text=white,
      inner sep=0pt,
      minimum size=1.1em,
      font=\bfseries\footnotesize,
      align=center] (char) {#1};
  }%
}

\definecolor{mygreen}{RGB}{34,139,34} 
\definecolor{myred}{RGB}{205,92,92}   

\newcommand{\cmark}{\textcolor{mygreen}{\ding{51}}} 

\usepackage{colortbl}
\definecolor{tablegray}{RGB}{230, 230, 230}

\DeclareRobustCommand{\GreenStep}[1]{\greenstep{#1}}

\newcommand{\teaserblock}{%
  \par\vspace{0.06in}%
  {\normalfont\normalsize
  \setcounter{figure}{0}%
  \refstepcounter{figure}\label{fig:teaser}%
  \begin{minipage}{0.96\textwidth}
  \centering
  \includegraphics[width=\linewidth]{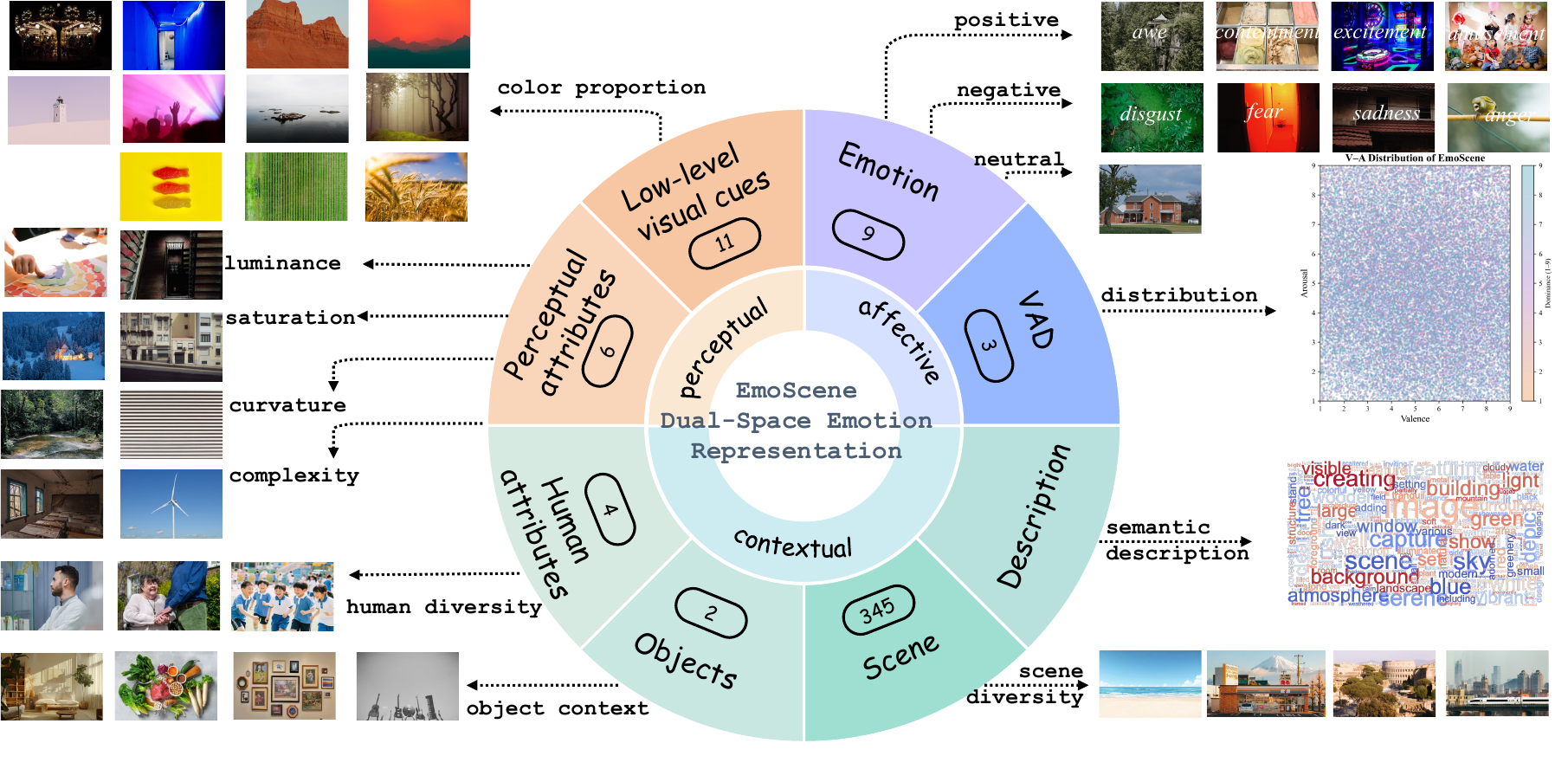}\\[-0.4ex]
  \parbox{\linewidth}{\raggedright\small Figure~\thefigure: \textbf{Dual-space annotations in EmoScene:} EmoScene represents visual affect in two complementary spaces. The affective space comprises discrete emotions and their VAD distribution; the perceptual space comprises color proportion, luminance, saturation, curvature, and visual complexity. Contextual annotations of people, objects, scenes, and text provide semantic grounding for both spaces. The total count of annotated attributes is shown in \ovalbox{circled boxes}, together with representative examples from each annotation group.}
  \end{minipage}}%
}

\makeatletter
\def\@maketitle{%
  \def\theauthors{\@author}
  \newcounter{eqfn}\setcounter{eqfn}{0}%
  \newsavebox{\titlearea}
  \sbox{\titlearea}{
    \let\footnote\relax\let\thanks\relax%
    \setcounter{footnote}{0}%
    \def\equalcontrib{%
      \ifnum\value{eqfn}=0%
        \footnote{These authors contributed equally.}%
        \setcounter{eqfn}{\value{footnote}}%
      \else%
        \footnotemark[\value{eqfn}]%
      \fi%
    }%
    \vbox{%
      \hsize\textwidth%
      \linewidth\hsize%
      \vskip 0.625in minus 0.125in%
      \centering%
      {\LARGE\bf \@title \par}%
      \vskip 0.1in plus 0.5fil minus 0.05in%
      {\Large{\textbf{\theauthors\ifhmode\\\fi}}}%
      \vskip .2em plus 0.25fil%
      {\normalsize \aaai@affiliations\ifhmode\\\fi}%
      \teaserblock%
      \vskip 1em plus 2fil%
    }%
  }%
  \newlength\actualheight%
  \settoheight{\actualheight}{\usebox{\titlearea}}%
  \ifdim\actualheight>\titlebox%
    \setlength{\titlebox}{\actualheight}%
  \fi%
  \vbox to \titlebox {%
    \let\footnote\thanks\relax%
    \setcounter{footnote}{0}%
    \def\equalcontrib{%
      \ifnum\value{eqfn}=0%
        \footnote{These authors contributed equally.}%
        \setcounter{eqfn}{\value{footnote}}%
      \else%
        \footnotemark[\value{eqfn}]%
      \fi%
    }%
    \hsize\textwidth%
    \linewidth\hsize%
    \vskip 0.625in minus 0.125in%
    \centering%
    {\LARGE\bf \@title \par}%
    \vskip 0.1in plus 0.5fil minus 0.05in%
    {\Large{\textbf{\theauthors\ifhmode\\\fi}}}%
    \vskip .2em plus 0.25fil%
    {\normalsize \aaai@affiliations\ifhmode\\\fi}%
    \teaserblock%
    \vskip 1em plus 2fil%
  }%
}%
\makeatother

\title{EmoScene: A Dual-Space Dataset for Controllable Affective Image Generation}
\author{
    Li He\textsuperscript{\rm 1},
    Longtai Zhang\textsuperscript{\rm 1},
    Wenqiang Zhang\textsuperscript{\rm 1},
    Yan Wang\textsuperscript{\rm 2},
    Lizhe Qi\textsuperscript{\rm 1}
}
\affiliations{
    \textsuperscript{\rm 1}Fudan University\\
    \textsuperscript{\rm 2}East China Normal University\\
    \{23110860006, 25213070120\}@m.fudan.edu.cn, \{wqzhang, qilizhe\}@fudan.edu.cn\\
    yanwang@dase.ecnu.edu.cn
}

\begin{document}

\maketitle
\setcounter{figure}{1}
\begin{abstract}
Text-to-image diffusion models achieve high visual fidelity, yet fine-grained affective control remains difficult because textual emotion cues often fail to specify the visual perceptual factors underlying affective expression. Existing visual-affect datasets are likewise often limited to discrete labels, specific domains, or limited supervision of perceptual attributes. We introduce EmoScene, a large-scale dual-space dataset for controllable affective image generation, containing 1.2M images across more than 300 scene categories. Its affective space jointly represents discrete emotions and continuous valence--arousal--dominance (VAD), its perceptual space records measurable appearance attributes, and contextual descriptions ground both in scene semantics. EmoScene combines multi-model annotation with human-in-the-loop quality control. A random audit of 30,519 images yields 91.22\% agreement on discrete emotion labels, while an independent multi-rater evaluation yields Fleiss' $\kappa=0.85$. After controlling for source and scene composition, affective dimensions and perceptual attributes exhibit stable associations across data sources, reflecting statistical tendencies rather than deterministic visual rules. To demonstrate the dataset's utility, we further develop AffectCtrl, which learns residuals in the conditioning space of frozen diffusion models to support categorical emotion generation and continuous control over VAD, brightness, and saturation. AffectCtrl achieves 85.75\% categorical emotion accuracy, outperforms EmotiCrafter in valence and arousal control under a shared evaluation protocol, and obtains Pearson correlations of 0.673--0.765 across all five continuous axes. These results demonstrate that EmoScene provides a scalable data foundation for analyzing and controlling affective expression in visual generation.

\end{abstract}
\section{Introduction}
\label{sec:intro}

\begin{center}
    \textit{``To create emotion is to organize perception.''}
\end{center}
\hfill \textit{--inspired by Rudolf Arnheim\cite{arnheim1954art, arnheim2023visual}}

Visual affect depends not only on depicted content but also on perceptual organization. Although affective computing has advanced emotion recognition and reasoning~\cite{afzal2024comprehensive,ma2025review,lian2024gpt}, current generative systems still offer limited control over how visual content feels.

Text-to-image diffusion models~\cite{chen2024pixart,balaji2022ediff,saharia2022photorealistic,rombach2022high} primarily control affect through prompt modifiers, such as changing ``happy'' to ``extremely happy.'' Such wording does not specify the perceptual carriers of affect, including luminance and color. Existing affective datasets and generators likewise provide limited supervision for continuous intensity and appearance-level expression~\cite{yang2023emoset,yang2024emogen,yuan2026coemogen,dang2025emoticrafter}.

The central bottleneck is joint supervision: existing datasets rarely align categorical emotion, continuous Valence--Arousal--Dominance (VAD), measurable perceptual attributes, and contextual semantics across diverse scenes. Without such data, models cannot reliably learn affect--perception associations or expose interpretable controls beyond textual modifiers.

We introduce EmoScene, a large-scale dual-space dataset containing 1.2M images across more than 300 scene categories. Here, dual-space denotes two complementary representations of visual affect: an affective space of discrete emotions and continuous VAD, and a perceptual space of measurable appearance attributes. Contextual descriptions ground both spaces in scene semantics and support scene-aware stratification. To demonstrate EmoScene's utility, we develop AffectCtrl, a lightweight reference controller that learns conditioning-space residuals over frozen diffusion generators. AffectCtrl supports categorical emotion generation and continuous VAD, brightness, and saturation control, improving EmoGen-style categorical control and EmotiCrafter-style VA conditioning.

In summary, our contributions are:
\begin{itemize}
\item We construct EmoScene, a 1.2M-image dual-space dataset that aligns affective representations (discrete emotion and VAD) with measurable perceptual attributes, grounded in diverse scene contexts.
\item We provide a large-scale, scene-aware characterization of the statistical coupling between categorical emotions, continuous VAD dimensions, and measurable perceptual attributes in natural imagery.
\item We develop AffectCtrl, a lightweight dataset-grounded reference controller that demonstrates EmoScene's utility for categorical emotion generation and continuous control over VAD, brightness, and saturation.
\end{itemize}

\begin{figure}[!t]
  \centering
  \includegraphics[width=\linewidth,trim=0 2 4 3,clip]{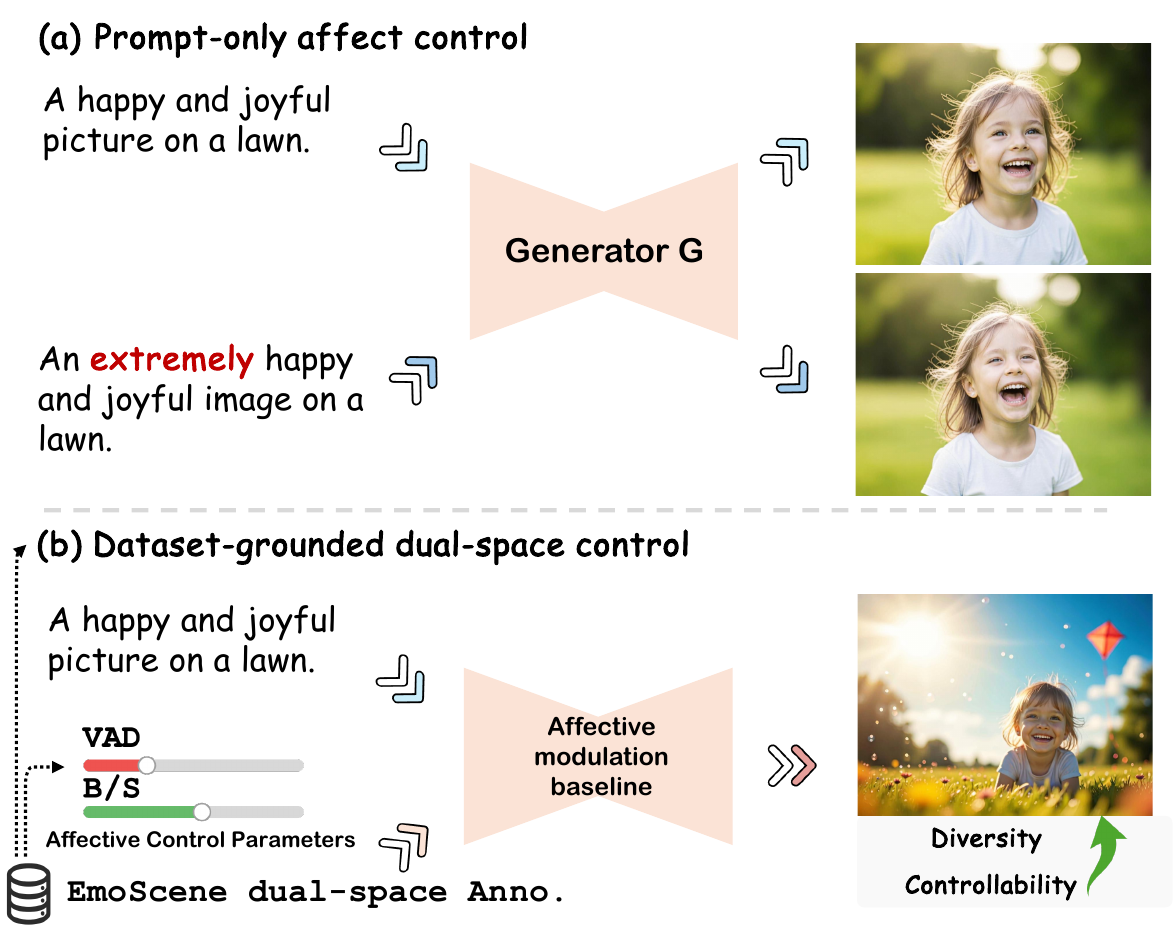}
  \vspace{-3mm}
  \caption{\textbf{Motivation.} EmoScene provides explicit VAD and brightness/saturation controls beyond underspecified emotion prompts.}
  \vspace{-3.5mm}
  \label{fig:motivation}
\end{figure}
\section{Related Work}
\label{sec:Related Work}

Our work is most closely related to two lines of work: affective visual datasets for generation and affect-controllable image synthesis.
\begin{table*}[!t]
\centering
\resizebox{\textwidth}{!}{%
\begin{tabular}{lccccccccc}
\toprule
\multicolumn{3}{c}{\textbf{General Info}} &
\multicolumn{2}{c}{\textbf{Affective Space}} &
\multicolumn{1}{c}{\textbf{Perceptual Space}} &
\multicolumn{3}{c}{\textbf{Contextual Semantics}} \\
\cmidrule(lr){1-3} \cmidrule(lr){4-5} \cmidrule(lr){6-6} \cmidrule(lr){7-9}
\textbf{Dataset} & \textbf{Year} & \textbf{\# Images} & \textbf{Emotion Type} & \textbf{VAD} & \textbf{Perceptual Attributes} & \textbf{Subject-level Anno.} & \textbf{Scene Anno.} & \textbf{Text/Lang. Anno.} \\
\midrule
ArtPhoto\cite{ArtPhoto} & 2010 & 0.8k & 8 & - & - & - & - & -  \\
Emotion6\cite{emotion6} & 2015 & 1.9k & 6+1 & V/A & - & - & - & -  \\
FI\cite{FI} & 2016 & 23k  & 8 & - & - & - & - & -  \\
IESN\cite{zhao2016IESN} & 2016 & 1000K & 8 & V/A/D & - & - & \cmark & \cmark  \\
Emotic\cite{kosti2017emotic} & 2017 & 18k & 26 & V/A/D & - & \cmark (human) & \cmark & -  \\
AffectNet\cite{mollahosseini2017affectnet} & 2017 & 450k & 8+3 & V/A & - & \cmark (face) & - & - \\
OASIS\cite{kurdi2017OASIS} & 2017 & 900 & - & V/A & - & - & \cmark & - \\
Artemis\cite{achlioptas2021artemis} & 2021 & 81k& 8+1 & - & - & - & \cmark & \cmark \\
HECO\cite{yang2022HECO} & 2022 & 9k& 8 & V/A/D & - & \cmark (human) & \cmark & - \\
Emoset\cite{yang2023emoset} & 2023 & 118k& 8 & - & \cmark(luminance, chroma) & \cmark (object) & \cmark & - \\
FindingEmo\cite{mertens2024findingemo} & 2024 & 26k & 24 & V/A & - & \cmark (multi-person) & \cmark & - \\
EmotiCrafter\cite{dang2025emoticrafter} & 2025 & 39k & - & V/A & - & - & - & \cmark \\
\midrule
\rowcolor{tablegray}\textbf{Ours} & \textbf{2026} & \textbf{1200K} & \textbf{8+1} & \textbf{\cmark (V/A/D)} & \textbf{\cmark (color, curvature, etc.)} & \textbf{\cmark (human, object)} & \textbf{\cmark} & \textbf{\cmark}  \\
\bottomrule
\end{tabular}%
}
\caption{Comparison of representative emotion-related datasets and our proposed EmoScene. Columns distinguish its affective and perceptual representation spaces from the contextual annotations that ground them, while highlighting the generation-oriented design of the dataset. 
Abbreviations: 
\#Images = number of labeled images; 
\textit{Perceptual Attributes} include color, luminance, curvature, and complexity; 
\textit{Subject-level Anno.} refers to annotations for specific entities(e.g., human or object); 
\textit{Text/Lang. Anno.} indicates the presence of textual or language-level descriptions (e.g.,prompts).}
\label{tab:dataset}
\end{table*}
\subsection{Affective Visual Datasets for Generation}
Existing affective visual datasets cover complementary portions of the required supervision. Early benchmarks emphasize recognition from categorical emotions, distributions, or VA ratings~\cite{machajdik2010affective,ArtPhoto,emotion6,FI,lian2021ctnet,lian2024merbench,fang2025emoe}, while EmoSet~\cite{yang2023emoset} adds perceptual and scene-level attributes. Emotic~\cite{kosti2017emotic}, AffectNet~\cite{mollahosseini2017affectnet}, IESN~\cite{zhao2016IESN}, and HECO~\cite{yang2022HECO} provide dimensional affect or social context; OASIS~\cite{kurdi2017OASIS} and FindingEmo~\cite{mertens2024findingemo} target full scenes; and Artemis~\cite{achlioptas2021artemis} and EmotiCrafter~\cite{dang2025emoticrafter} add language-level supervision. As Table~\ref{tab:dataset} shows, these datasets do not jointly align categorical affect, VAD, contextual semantics, and perceptual attributes across diverse scenes.
\subsection{Affective Image Generation and Editing}
Early affective image generation methods condition GANs on emotion labels or style vectors, as in AffectiveGAN~\cite{zhang2025affective,paskaleva2024unified,ijcai2018p780,wang2024surveyfacialexpressionrecognition,galanos2021affectgan}.

Affective diffusion methods use learned tokens, adapters, or numeric conditioning. EmoGen~\cite{yang2024emogen} learns a CLIP-aligned affective space, EmoEdit~\cite{yang2025emoedit} edits affect through a plug-and-play adapter, and EmotiCrafter~\cite{dang2025emoticrafter} injects continuous VA conditions. CoEmoGen~\cite{yuan2026coemogen} improves discrete emotional image generation through sentence-level captions and hierarchical LoRA; in contrast, our dataset-level representation additionally exposes VAD and perceptual variables. EmoCtrl~\cite{yang2025emoctrl} and EmoSpace~\cite{wang2026emospace} strengthen discrete emotional semantics, while AttriCtrl~\cite{chen2025attrictrl} controls individual aesthetic attributes with numeric instructions. These methods address categorical affect, semantic coherence, or isolated continuous axes, whereas EmoScene aligns categorical affect, VAD, perceptual attributes, and context within one generation-oriented dataset.
\begin{figure*}[!t]
  \centering
   \includegraphics[width=\textwidth]{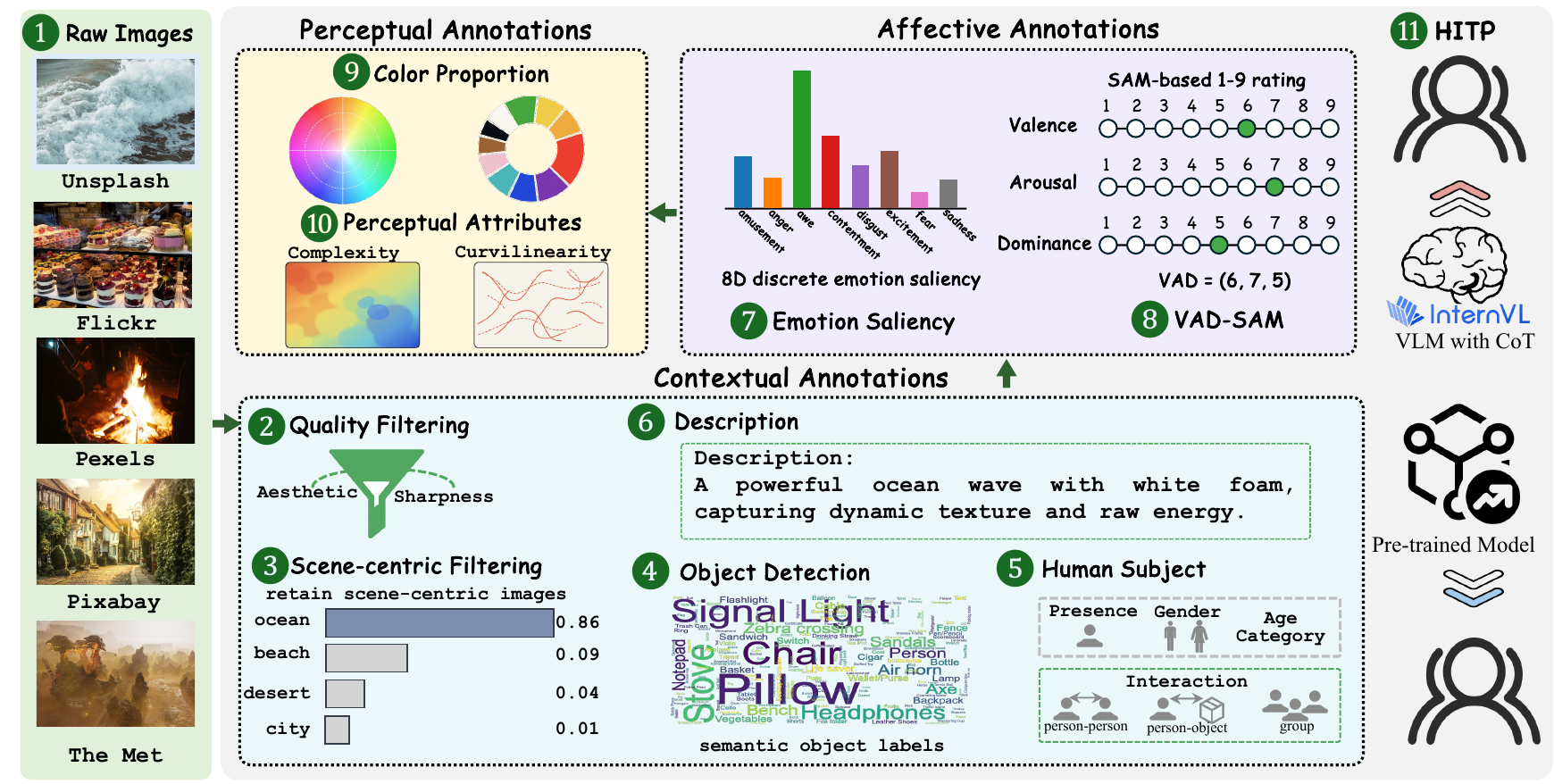}
   \vspace{-3mm}
   \caption{
   \textbf{Dual-Space Annotation Pipeline.}
EmoScene proceeds through image curation, contextual-semantic parsing, affective and perceptual annotation, and human-in-the-loop refinement. Full implementation details are provided in the supplementary material.}
   \vspace{-3.5mm}
   \label{fig:onecol}
\end{figure*}
\section{EmoScene Dataset}
EmoScene contains 1.2M images from more than 300 scene categories spanning everyday photographs, social interactions, and artworks. Each image includes affective-space labels (discrete emotion and continuous VAD), perceptual-space descriptors, and contextual text that grounds both, jointly supporting affect--perception analysis and controllable image generation. Source-specific licensing, privacy safeguards, and release and opt-out policies are documented in Supplementary Sec.~S2.4.
\subsection{Dual-space annotation pipeline}
\noindent\textbf{Image curation.}
Steps~\GreenStep{1}--\GreenStep{3} assemble diverse photographic and artistic scenes, remove low-quality samples using aesthetic and sharpness checks, and verify scene labels and image--text consistency with Places365 and CLIP. This produces a visually sound and semantically coherent candidate pool.

\noindent\textbf{Contextual semantics.}
Steps~\GreenStep{4}--\GreenStep{6} identify people, objects, attributes, and interactions, including age groups defined by the JRDB-Social taxonomy~\cite{jahangard2024jrdb}. Human-Aware Modeling~\cite{jiang2025modeling} and a multimodal language model then convert these relations into short captions and fine-grained contextual descriptions.

\noindent\textbf{Affective and perceptual spaces.}
Steps~\GreenStep{7}--\GreenStep{8} use Qwen2.5 and InternVL3-8B~\cite{zhu2025internvl3} to annotate eight emotions plus Neutral and VAD scores, with outputs aggregated through a human-aligned protocol. Steps~\GreenStep{9}--\GreenStep{10} complement this affective space with deterministic color and structural measurements in the perceptual space.
Using two affect-annotation models reduces reliance on the calibration of any single model, while deterministic perceptual measurements provide reproducible quantities rather than additional subjective labels. This separation lets EmoScene connect semantic judgments to measurable image properties without treating either as a proxy for the other.

\noindent\textbf{Human-in-the-loop quality control.}
In Step~\GreenStep{11}, trained annotators prioritize model disagreements and intensity mismatches, then feed corrections back into the annotation pipeline. The supplementary material details all models, prompts, feature definitions, and review procedures.

\begin{figure}[!t]
    \centering
    \includegraphics[width=\linewidth]{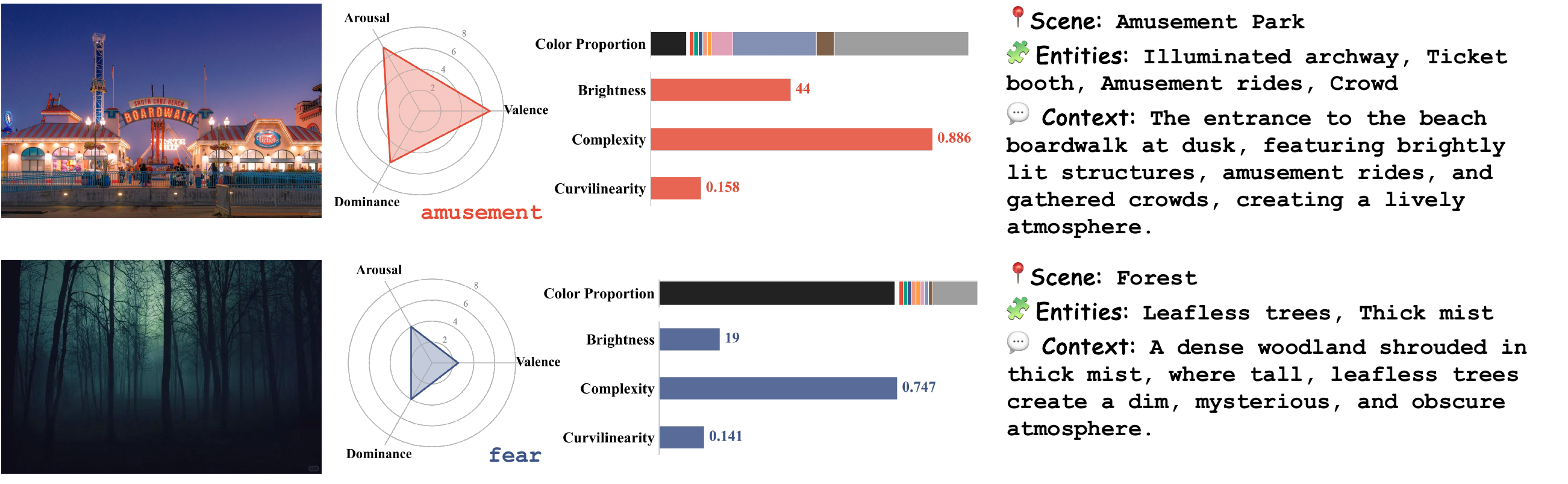} 
    \vspace{-3mm}
    \caption{\textbf{Representative examples of EmoScene's dual-space annotations.} }
    \vspace{-3.5mm}
    \label{fig:annotation_hero}
\end{figure}

\noindent\textbf{Representative annotation examples.} Figure~\ref{fig:annotation_hero} contrasts an amusement park and a mist forest. Their VAD labels co-vary with brightness, complexity, and semantic context, illustrating the annotation scheme without implying deterministic visual rules.

\begin{figure}[!t]
  \centering
  \includegraphics[width=\linewidth]{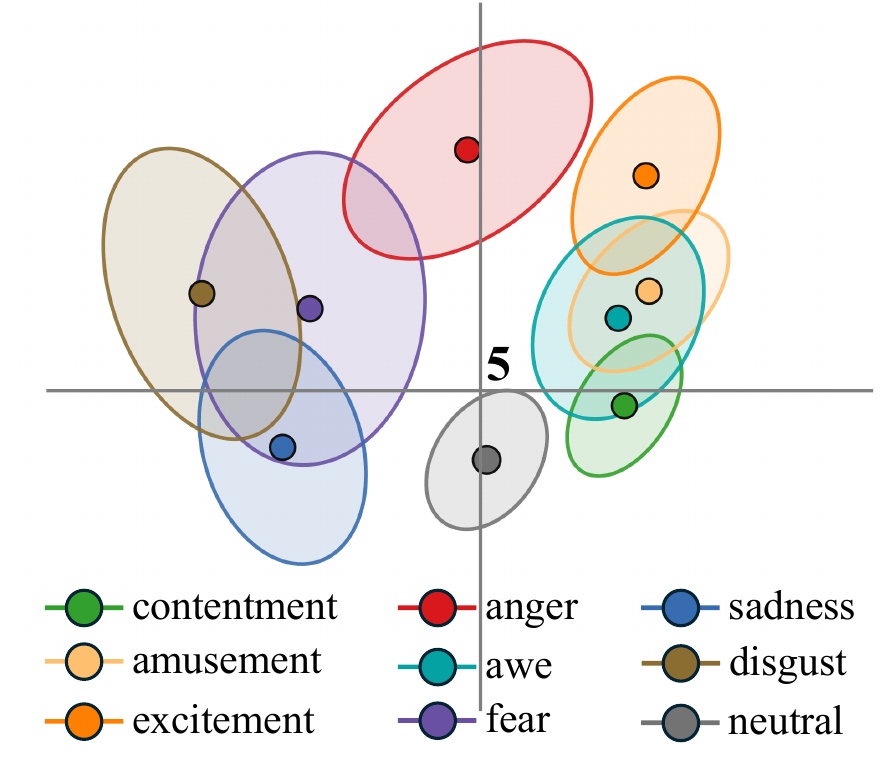}
  \vspace{-3mm}
  \caption{\textbf{Valence--Arousal structure of EmoScene.} Each emotion is represented by its class-normalized centroid and covariance ellipse in the VA plane. The overlap between ellipses reflects shared affective regions and label ambiguity, while the global layout reveals a circumplex-like organization across valence and arousal.}
  \label{fig:affective_structure}
  \vspace{-1.5mm}
  \includegraphics[width=\linewidth,trim=0 14bp 0 16bp,clip]{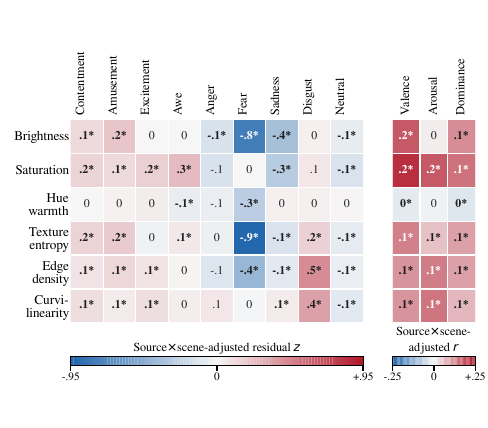}
  \vspace{-3mm}
  \caption{\textbf{Source$\times$scene-adjusted affect--perception coupling.} Left: category-wise mean standardized perceptual residuals ($N=18{,}085$). Right: Pearson correlations between perceptual features and VAD dimensions ($N=27{,}252$). Asterisks require a 95\% scene-cluster bootstrap interval excluding zero and BH-adjusted $q<.05$; values are rounded to one decimal for column-scale readability.}
  \vspace{-3.5mm}
  \label{fig:scene_aware_coupling}
\end{figure}

\subsection{Dataset Analysis and Statistics}
\noindent\textbf{Affective structure.}
We first ask whether the discrete labels form a coherent geometry in the continuous Valence--Arousal (VA) plane. Figure~\ref{fig:affective_structure} summarizes each emotion by its class-normalized centroid and covariance ellipse, so the display reflects within-class variation rather than raw class frequency. Contentment, amusement, excitement, and awe occupy the high-valence side, whereas fear, sadness, and disgust shift toward low valence; excitement separates from contentment chiefly along arousal. Neutral remains near the center at lower arousal. The overlap between neighboring ellipses is also informative: the labels occupy continuous affective regions with shared ambiguity, motivating VA as a bridge from discrete categories to perceptual analysis.

\noindent\textbf{Scene-aware affect--perception coupling.}
Raw visual differences can be confounded by dataset composition: an emotion may occur disproportionately in a particular source or scene. We therefore analyze a fixed, source--scene-stratified cohort and subtract the corresponding source$\times$scene mean from each perceptual feature. The left panel of Fig.~\ref{fig:scene_aware_coupling} reports the resulting category-wise standardized residuals. Fear is markedly darker ($-0.80$) and less textured ($-0.94$), sadness is darker ($-0.36$) and less saturated ($-0.28$), and disgust has denser edges ($+0.46$) and greater curvilinearity ($+0.37$). In contrast, contentment and amusement exhibit broadly positive brightness, saturation, and texture profiles.

The right panel tests whether these class-level profiles also align with continuous affect after removing the same source$\times$scene variation. Saturation has the strongest associations with valence ($r=0.22$) and arousal ($r=0.18$); brightness tracks valence ($r=0.18$) but is nearly unrelated to arousal ($r=0.01$). Arousal additionally aligns with curvilinearity ($r=0.15$) and edge density ($r=0.14$), whereas dominance has smaller, distributed associations rather than a single dominant low-level cue. Together, Figs.~\ref{fig:affective_structure} and~\ref{fig:scene_aware_coupling} characterize the scene-aware statistical coupling across categorical emotion, continuous affect, and measurable perceptual attributes---the second contribution of EmoScene. These controlled associations do not establish causal visual mechanisms, but they show that the annotations carry complementary, reproducible structure for joint analysis and control. The principal associations remain stable across raw, scene-adjusted, and source$\times$scene-adjusted analyses (Supplementary Fig.~S7).

\subsection{Annotation Reliability and Validation}
To evaluate the effectiveness of our HITL pipeline, we conducted a rigorous validation on three fronts, as summarized in Table~\ref{tab:hitl}. 

\noindent\textbf{Large-scale Audit.} We performed a random audit of 30,519 images (2.54\% of the total), achieving a high agreement of 91.22\% on discrete emotions and remarkably low Mean Squared Errors (MSE) in the VAD space. An independent multi-rater evaluation yielded a Fleiss' $\kappa$ of 0.85, indicating strong agreement on the audited subset. Annotators verify each annotation field with independent Yes/No checks; conflicts are resolved via evidence-based adjudication and rejected items enter a second-pass refinement loop (see Supp.Sec.~2.3).

\noindent\textbf{Bias Mitigation Stress-test.} To ensure the dataset captures deep semantics rather than shallow visual shortcuts (e.g., darkness equals sadness), we audited 297 counter-intuitive scenes (e.g., candle-lit dinners with low-luminance but high-pleasantness). Compelling visual proofs of these bias-mitigation examples are extensively showcased in the supplementary material. Human annotators maintained an 85.0\% agreement with MLLM labels in these challenging cases, confirming the semantic depth of EmoScene.

\begin{table}[!t]
  \setcounter{table}{1}
  \centering
  \resizebox{\columnwidth}{!}{%
  \begin{tabular}{lcc}
    \hline
    \rowcolor{tablegray}
    \textbf{Subset} & \textbf{Discrete (Acc./ $\kappa$)} & \textbf{VAD (MSE / $r$ )} \\
    \hline
    Random (30,519) & 91.22\%(Acc.) & 0.070 / 0.050 / 0.033(MSE) \\
    Bias-test (297) & 85.00\%(Acc.) & 0.181 / 0.102 / 0.084(MSE) \\
    Multi-rater & 0.85 ($\kappa$) & 0.960 / 0.984 / 0.901($r$) \\
    \hline
  \end{tabular}
  }
  \vspace{-3mm}
  \caption{Human verification of MLLM labels (HITL Audit). We report the agreement between MLLM-generated labels and human ground truth across three subsets. Our labels achieve ``almost perfect'' agreement ($\kappa=0.85$) and extremely low VAD error.}
  \vspace{-3.5mm}
  \label{tab:hitl}
\end{table}

\section{Method}

\subsection{Overview: Dual-Space Affect Control}
We instantiate EmoScene's two spaces as complementary interfaces in \emph{AffectCtrl}: categorical emotion and continuous VAD specify affective targets, while brightness and saturation provide representative perceptual controls. AffectCtrl is a lightweight reference controller rather than a new diffusion architecture. Both branches learn additive residuals in the conditioning space of a frozen text-to-image generator.

\begin{figure}[!t]
  \centering
   \includegraphics[width=\linewidth]{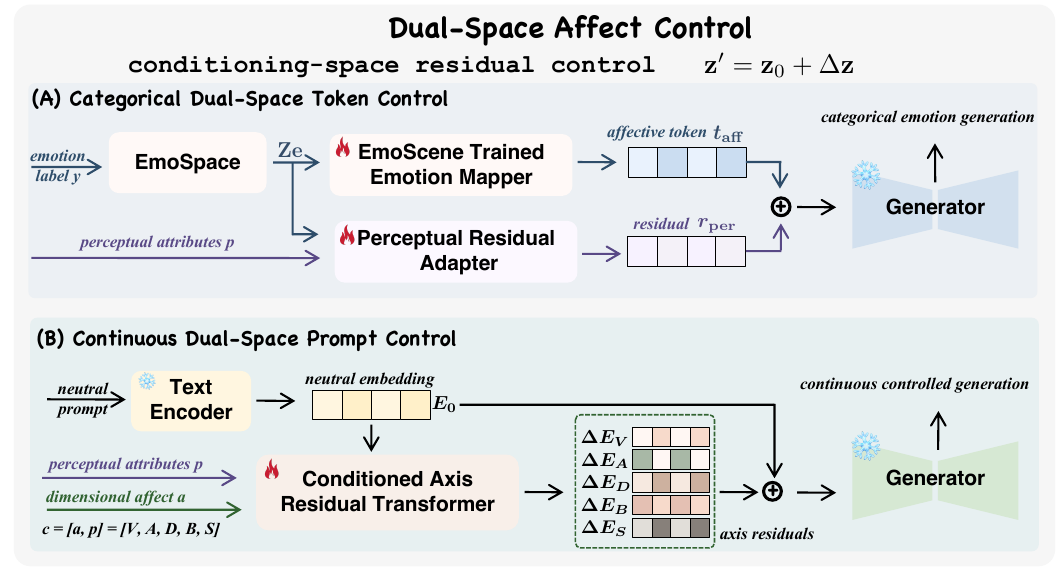}
   \vspace{-3mm}
   \caption{
    \textbf{Overview of AffectCtrl.}
The categorical branch maps an EmoSpace embedding to an affective token and adds a brightness--saturation residual. The continuous branch adds axis-wise VAD/brightness/saturation residuals to neutral SDXL embeddings. Both generators remain frozen.
}
   \vspace{-3.5mm}
   \label{fig4}
\end{figure}

Let $\mathbf{Z}_0\in\mathbb{R}^{m\times d}$ denote the base conditioning, with one affective token ($m=1$) in the categorical branch and an SDXL prompt-embedding matrix in the continuous branch. The controls are a categorical code $\mathbf{c}_{\mathrm{cat}}\in\{0,1\}^{K}$, VAD vector $\mathbf{c}_{\mathrm{vad}}\in\mathbb{R}^{3}$, and brightness--saturation vector $\mathbf{c}_{\mathrm{per}}\in\mathbb{R}^{2}$. AffectCtrl predicts
\begin{equation}
\begin{aligned}
\mathbf{Z}' &= \mathbf{Z}_0 + \Delta \mathbf{Z},\\
\Delta \mathbf{Z}
&= R(\mathbf{Z}_0,\mathbf{c}_{\mathrm{cat}},
\mathbf{c}_{\mathrm{vad}},\mathbf{c}_{\mathrm{per}};\theta),\\
\Delta \mathbf{Z} &\in \mathbb{R}^{m\times d}.
\end{aligned}
\label{eq:unified_residual}
\end{equation}
where $R(\cdot;\theta)$ is trainable and $\Delta\mathbf{Z}$ is the control residual. The categorical and continuous branches instantiate $R$ for an EmoGen-style affective token and SDXL prompt embeddings, respectively (Fig.~\ref{fig4}).

\subsection{EmoScene-Grounded Control Supervision}
EmoScene provides categorical labels, continuous VAD, and HSV-derived brightness and saturation, separating the requested affect from its visual realization.

The categorical branch uses an EmoGen-compatible split balanced over emotions, objects, and scenes. The continuous branch uses locked-prefix pairs: a neutral prompt and a target prompt with the same semantic prefix plus affective--perceptual cues. Their precomputed SDXL embedding difference therefore supervises control displacement while limiting content replacement.

\subsection{Categorical Dual-Space Token Control}
Given an emotion $y\in\{1,\ldots,K\}$, we retrieve a fixed EmoSpace prototype or sampled embedding $\mathbf{z}_e\in\mathbb{R}^{d_e}$ and train an EmoScene mapper $M_{\mathrm{cat}}:\mathbb{R}^{d_e}\rightarrow\mathbb{R}^{d_t}$ for the frozen EmoGen-style generator~\cite{yang2024emogen}:
\begin{equation}
\mathbf{t}_{\mathrm{aff}} = M_{\mathrm{cat}}(\mathbf{z}_e;\theta_{\mathrm{cat}}),
\qquad
\mathbf{t}_{\mathrm{aff}}\in\mathbb{R}^{d_t}.
\label{eq:cat_token}
\end{equation}
Only the mapper parameters $\theta_{\mathrm{cat}}$ are learned, isolating EmoScene's categorical supervision under the matched interface.

We then freeze $M_{\mathrm{cat}}$ and condition a perceptual adapter $M_{\mathrm{per}}:\mathbb{R}^{d_e+2}\rightarrow\mathbb{R}^{d_t}$ on brightness and saturation $\mathbf{p}=[b,s]^\top$:
\begin{equation}
\mathbf{r}_{\mathrm{per}} = M_{\mathrm{per}}(\mathbf{z}_e,\mathbf{p};\theta_{\mathrm{per}}),
\qquad
\mathbf{t}_{\mathrm{final}} = \mathbf{t}_{\mathrm{aff}} + \lambda_{\mathrm{per}}\mathbf{r}_{\mathrm{per}}.
\label{eq:cat_residual}
\end{equation}
The scale $\lambda_{\mathrm{per}}$ controls refinement; $\mathbf{t}_{\mathrm{final}}$ replaces the emotion placeholder. The additive form preserves the categorical token while modifying its appearance.

\subsection{Continuous Dual-Space Prompt Control}
The continuous branch receives a neutral prompt $x$ and
\begin{equation}
\mathbf{c} = [V,A,D,b,s]^\top\in\mathbb{R}^{5},
\end{equation}
where VAD denotes dimensional affect and $(b,s)$ denotes perceptual control. A frozen SDXL text encoder produces token embeddings $\mathbf{E}_0\in\mathbb{R}^{n\times d_s}$ and pooled embedding $\mathbf{g}_0\in\mathbb{R}^{d_g}$. For $I=\{V,A,D,b,s\}$, a lightweight transformer predicts axis-wise token residuals and one pooled residual:
\begin{equation}
\begin{aligned}
\{\Delta \mathbf{E}_j\}_{j\in I},\,\Delta \mathbf{g}
&= R_{\mathrm{cont}}(\mathbf{E}_0,\mathbf{g}_0,
\mathbf{c};\theta_{\mathrm{cont}}),\\
\Delta \mathbf{E}_j &\in\mathbb{R}^{n\times d_s},\quad
\Delta \mathbf{g}\in\mathbb{R}^{d_g}.
\end{aligned}
\label{eq:axis_residual}
\end{equation}
The controlled conditioning is formed additively in the same embedding spaces:
\begin{equation}
\mathbf{E}_{\mathrm{ctrl}} = \mathbf{E}_0 + \sum_{j\in I} \Delta \mathbf{E}_j,
\qquad
\mathbf{g}_{\mathrm{ctrl}} = \mathbf{g}_0 + \Delta \mathbf{g}.
\label{eq:prompt_residual}
\end{equation}
Each $\Delta\mathbf{E}_j$ depends on the signed request $\mathbf{c}$; their sum provides named control directions from a fixed prompt. Frozen SDXL samples from $(\mathbf{E}_{\mathrm{ctrl}},\mathbf{g}_{\mathrm{ctrl}})$.

\subsection{Training and Inference}
\noindent\textbf{Categorical branch.}
We train $M_{\mathrm{cat}}$ with the EmoGen objective on the balanced EmoScene split while freezing the generator, EmoSpace encoder, and auxiliary networks. We then freeze $M_{\mathrm{cat}}$ and train $M_{\mathrm{per}}$ on brightness and saturation, separating categorical supervision from perceptual refinement.

\noindent\textbf{Continuous branch.}
For locked-prefix neutral and target embeddings $(\mathbf{E}_0,\mathbf{g}_0)$ and $(\mathbf{E}_t,\mathbf{g}_t)$, we optimize
\begin{equation}
L_{\mathrm{cont}}
= L_{\mathrm{tok}}
+ \lambda_p L_{\mathrm{pool}}
+ \lambda_a L_{\mathrm{axis}}
+ \lambda_0 L_{0}.
\label{eq:cont_loss}
\end{equation}
where $\lambda_p,\lambda_a,\lambda_0\geq0$ weight the three auxiliary terms. The reconstruction losses are
\begin{equation}
\begin{aligned}
L_{\mathrm{tok}} &=
\frac{1}{n d_s}\|\mathbf{E}_{\mathrm{ctrl}}-\mathbf{E}_t\|_{F}^{2},\\
L_{\mathrm{pool}} &=
\frac{1}{d_g}\|\mathbf{g}_{\mathrm{ctrl}}-\mathbf{g}_t\|_{2}^{2}.
\end{aligned}
\label{eq:loss_terms}
\end{equation}
$L_{\mathrm{axis}}$ regresses $\mathbf{c}$ from the transformer representation, while $L_0$ applies the reconstruction criterion at $\mathbf{c}=\mathbf{0}$ to recover neutral conditioning. Samples are weighted by EmoScene quality scores when available.

\noindent\textbf{Inference.}
Equations~\eqref{eq:cat_residual} and~\eqref{eq:prompt_residual} provide categorical and continuous conditioning to the respective frozen generators. Architectures, normalization, optimization, and inference details are provided in Supplementary Sec.~S4.


\section{Experiments}
\label{sec:experiments}

We evaluate AffectCtrl on categorical emotion control and continuous VAD, brightness, and saturation control, using EmoGen- and EmotiCrafter-style interfaces for matched comparisons.

\subsection{Experimental Setup}
\label{subsec:exp_setup}

Complete generation protocols, evaluator data provenance, and metric definitions, including ACS, are provided in Supplementary Secs.~S5.1--S5.3.

\paragraph{Categorical control.}
Under emotion-only control, we compare direct prompting and GPT-5.5 prompt rewriting with SDXL and FLUX.1, a CoEmoGen-style reproduction, and three EmoGen-style token controllers.  Learned methods generate 50 images per emotion with matched seeds and are assessed by common emotion and semantic evaluators.

\paragraph{Continuous control.}
We compare with EmotiCrafter on a shared $5\times5$ VA grid over 132 neutral prompts (3,300 images per method).  Direct and GPT-5.5-rewritten SDXL/FLUX.1 prompts test all five axes, while the released AttriCtrl (FLUX.1) checkpoint provides a brightness-control baseline.  Common VAD/HSV evaluators report errors on the normalized $[-3,3]$ scale.

\subsection{Categorical Dual-Space Token Control}
\label{subsec:cat_results}

Table~\ref{tab:cat_control} shows that learned controllers outperform prompt baselines. The CoEmoGen-style reproduction improves confidence and semantic consistency over EmoGen, but has slightly lower categorical accuracy (79.25\% versus 80.25\%). Retraining the same EmoGen mapper on EmoScene raises Emo-A from 80.25\% to 81.50\% and Sem-C from 0.603 to 0.726, isolating the benefit of dataset supervision under a matched controller. Adding the perceptual residual yields the best overall result, with 85.75\% Emo-A and 0.799 ACS.

\begin{table}[!t]
\centering
\scriptsize
\resizebox{\columnwidth}{!}{%
\begin{tabular}{lcccc}
\toprule
Method & Emo-A$\uparrow$ & Conf.$\uparrow$ & Sem-C$\uparrow$ & ACS$\uparrow$ \\
\midrule
SDXL direct prompt & 71.00 & 0.655 & 0.602 & 0.656 \\
FLUX.1 direct prompt & 73.25 & 0.672 & 0.577 & 0.661 \\
GPT-5.5 + SDXL & 71.75 & 0.657 & 0.677 & 0.684 \\
GPT-5.5 + FLUX.1 & 74.00 & 0.680 & 0.670 & 0.697 \\
CoEmoGen-style~\cite{yuan2026coemogen}$^{\dagger}$ & 79.25 & 0.722 & 0.669 & 0.728 \\
EmoGen~\cite{yang2024emogen} & 80.25 & 0.704 & 0.603 & 0.703 \\
EmoGen + EmoScene & 81.50 & 0.740 & 0.726 & 0.760 \\
\rowcolor{tablegray}
Ours & \textbf{85.75} & \textbf{0.789} & \textbf{0.750} & \textbf{0.799} \\
\bottomrule
\end{tabular}}
\vspace{-2.5mm}
\caption{Categorical generation under emotion-only control. $^{\dagger}$Paper-level CoEmoGen-style reproduction. Emo-A is reported as a percentage.}
\vspace{-3mm}
\label{tab:cat_control}
\end{table}

\begin{table}[!t]
\centering
\scriptsize
\setlength{\tabcolsep}{3pt}
\resizebox{\columnwidth}{!}{%
\begin{tabular}{lccccc}
\toprule
Method & Valence$\downarrow$ & Arousal$\downarrow$ & Dominance$\downarrow$ & Brightness$\downarrow$ & Saturation$\downarrow$ \\
\midrule
SDXL direct prompt & 1.905$\pm$1.398 & 1.849$\pm$1.453 & 2.041$\pm$1.396 & 1.623$\pm$0.923 & \textbf{1.360$\pm$0.908} \\
FLUX.1 direct prompt & 1.649$\pm$1.260 & 1.940$\pm$1.450 & 2.022$\pm$1.386 & 1.403$\pm$1.140 & 1.566$\pm$1.015 \\
GPT-5.5 + SDXL & 1.826$\pm$1.421 & \underline{1.689$\pm$1.403} & 1.958$\pm$1.407 & 1.566$\pm$0.886 & 1.569$\pm$1.075 \\
GPT-5.5 + FLUX.1 & 1.649$\pm$1.314 & 1.734$\pm$1.387 & \underline{1.903$\pm$1.340} & \underline{1.327$\pm$1.008} & 1.515$\pm$1.019 \\
EmotiCrafter~\cite{dang2025emoticrafter} & \underline{1.221$\pm$0.912} & 1.897$\pm$1.291 & N/A & N/A & N/A \\
AttriCtrl (FLUX.1)~\cite{chen2025attrictrl} & N/A & N/A & N/A & 1.914$\pm$1.317 & N/A \\
\rowcolor{tablegray}
Ours & \textbf{1.132$\pm$0.814} & \textbf{1.331$\pm$0.999} & \textbf{1.189$\pm$0.979} & \textbf{1.288$\pm$0.814} & \underline{1.412$\pm$1.176} \\
\bottomrule
\end{tabular}}
\vspace{-2.5mm}
\caption{Continuous control on 132 neutral prompts. EmotiCrafter supports VA only, AttriCtrl brightness only, and AffectCtrl the full VAD+perceptual interface. Errors use the normalized $[-3,3]$ scale.}
\vspace{-2.5mm}
\label{tab:continuous_control}
\end{table}

\subsection{Continuous Dual-Space Control}
\label{subsec:cont_results}

Table~\ref{tab:continuous_control} evaluates a shared VA interface and the extension to dominance, brightness, and saturation. Prompt baselines can follow literal brightness and saturation instructions, but remain less reliable for affective dimensions, especially dominance. AffectCtrl lowers valence/arousal error relative to EmotiCrafter (1.132/1.331 versus 1.221/1.897) and brightness error relative to AttriCtrl. VA is therefore the matched method comparison, whereas D/B/S evaluates the expanded interface enabled by EmoScene rather than a like-for-like capability shared by every baseline.

The MAE results are not uniformly best: direct SDXL prompting obtains lower saturation error (1.360 versus 1.412), despite weaker control on most other axes. Pearson correlations provide the complementary monotonicity test, confirming stronger VA control than EmotiCrafter (0.765/0.673 versus 0.701/0.138 for V/A) and clear target--response trends on D/B/S (0.756/0.762/0.718). Detailed baseline results are reported in the supplementary material.

To examine axis specificity, we measure all five responses while sweeping one requested control and fixing the remaining inputs at neutral values. Target-axis responses remain strong (mean diagonal $r=0.753$), although the controls are not fully orthogonal (mean absolute off-axis response $0.299$); the pooled and within-prompt matrices are reported in Supplementary Sec.~S5.4 and Fig.~S10.

\subsection{Human Evaluation}
\label{subsec:human_eval}

Table~\ref{tab:human_eval} reports 82.5\% categorical and 77.5\% VA pairwise wins, and 63.0\% acceptance for the complete interface; protocols are provided in the supplementary material.

\begin{table}[!t]
\centering
\small
\resizebox{\columnwidth}{!}{%
\begin{tabular}{lccc}
\toprule
Study & Protocol & Vote/Accept$\uparrow$ & Pair/Majority$\uparrow$ \\
\midrule
Ours vs. EmoGen & 160 pairs, 3 raters & 74.7\% & 82.5\% \\
Ours vs. EmotiCrafter & 80 pairs, 4 raters & 80.6\% & 77.5\% \\
Dual-space accept & 100 images, 4 raters & 67.8\% & 63.0\% \\
\bottomrule
\end{tabular}}
\vspace{-2.5mm}
\caption{Human evaluation of categorical, VA, and full dual-space controllability.}
\vspace{-3mm}
\label{tab:human_eval}
\end{table}

\subsection{Ablation and Analysis}
\label{subsec:ablation_analysis}

The categorical ablation in Table~\ref{tab:cat_control} shows that EmoScene supervision improves emotion and semantic scores, while the perceptual residual adapter raises ACS from 0.760 to 0.799. This supports modeling affective and perceptual controls as coupled spaces.

For the continuous branch, the matched Full model achieves the best macro correlation (0.757) and MAE (1.221) in Table~\ref{tab:continuous_ablation}. Removing $L_{\mathrm{pool}}$ is most harmful overall, whereas $L_{\mathrm{axis}}$ is especially important for arousal and dominance; per-axis results are provided in the supplementary material.

\begin{table}[!t]
\centering
\scriptsize
\begin{tabular*}{\columnwidth}{@{\extracolsep{\fill}}lcccc@{}}
\toprule
Metric & Full & w/o $L_{\mathrm{pool}}$ & w/o $L_{\mathrm{axis}}$ & w/o $L_0$ \\
\midrule
Avg. Corr.$\uparrow$ & \textbf{0.757} & 0.649 & 0.661 & 0.726 \\
Avg. MAE$\downarrow$ & \textbf{1.221} & 1.383 & 1.339 & 1.272 \\
\bottomrule
\end{tabular*}
\vspace{-2.5mm}
\caption{Matched continuous-objective ablation averaged over V/A/D/B/S. Full is retrained separately from the checkpoint in Table~\ref{tab:continuous_control}; MAE uses the normalized $[-3,3]$ scale.}
\vspace{-2.5mm}
\label{tab:continuous_ablation}
\end{table}

Arousal shows the largest shared-VA gain over EmotiCrafter. Dominance remains the hardest axis perceptually, but its automatic results indicate a usable control direction in the conditioning space.

\subsection{Qualitative Results}
\label{subsec:qualitative_analysis}

\begin{figure}[!t]
  \centering
  \includegraphics[width=\linewidth]{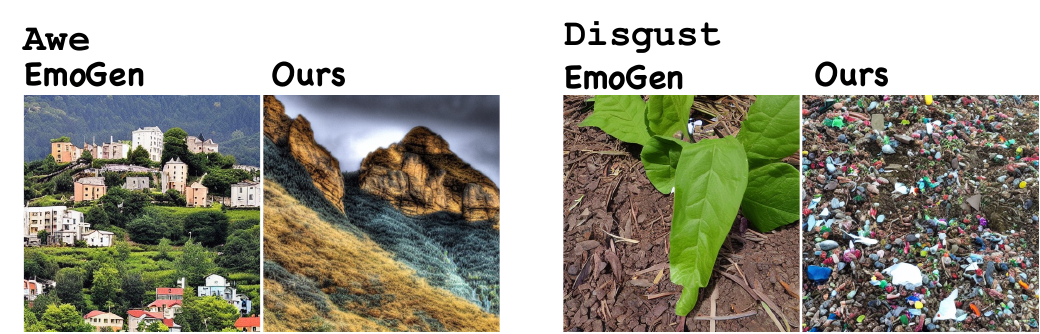}
  \vspace{-2.5mm}
  \caption{\textbf{Categorical comparison with EmoGen.} Under matched seeds and sampling, AffectCtrl produces more target-specific cues for awe and disgust.}
  \vspace{-2.5mm}
  \label{fig:categorical_qualitative}
\end{figure}

\begin{figure}[!t]
  \centering
  \includegraphics[width=0.92\linewidth]{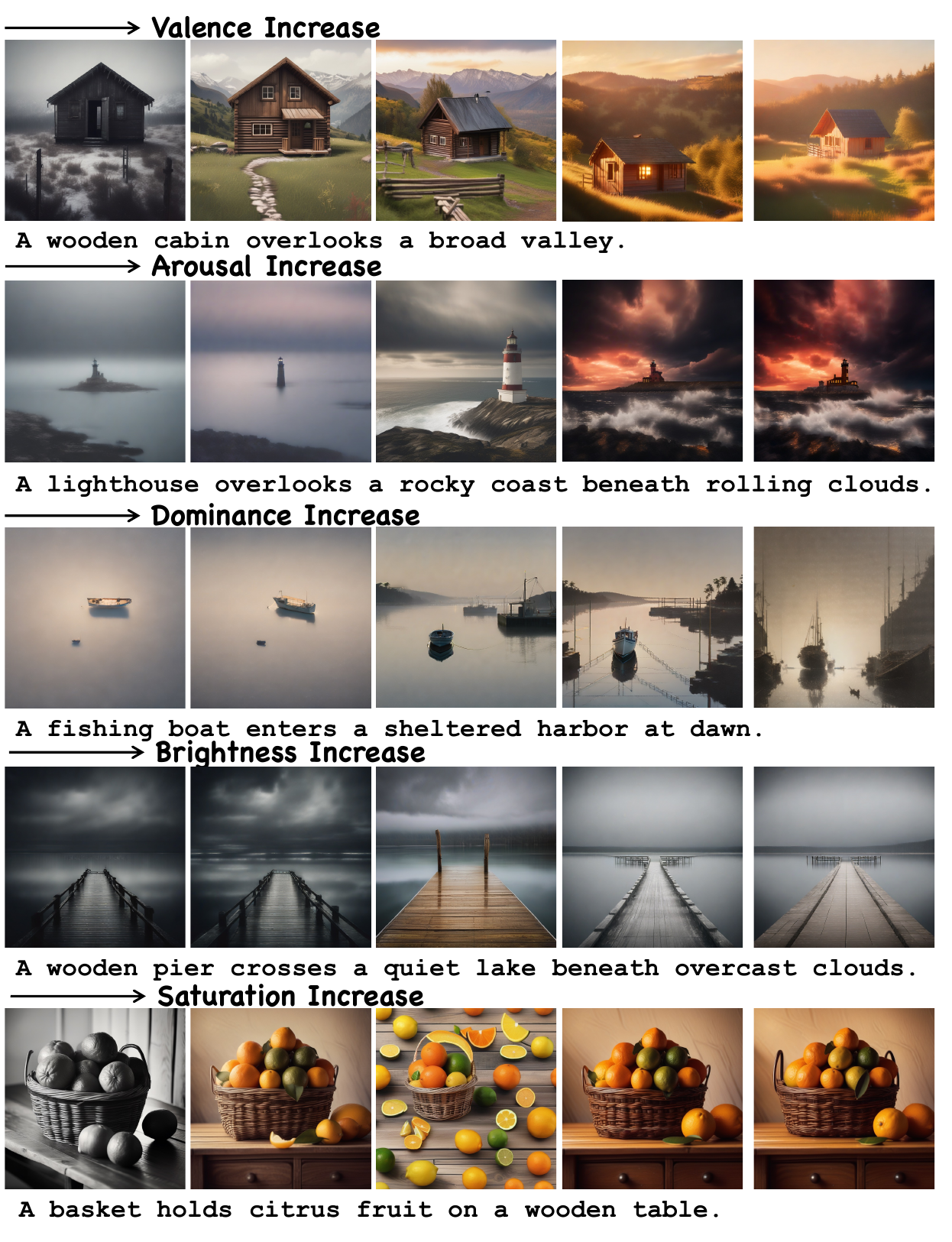}
  \vspace{-3.5mm}
  \caption{\textbf{Five-axis control with AffectCtrl.} Each row fixes the prompt and seed, fixes the other requested input controls at neutral values, and sweeps one axis from low to high, producing progressive changes while preserving recognizable content.}
  \vspace{-5mm}
  \label{fig:five_axis_qualitative}
\end{figure}

Figure~\ref{fig:categorical_qualitative} shows stronger target-specific cues than EmoGen for both awe and disgust. Fixed-prompt, fixed-seed sweeps in Fig.~\ref{fig:five_axis_qualitative} exhibit progressive changes along all five axes while retaining recognizable scene content.

\section{Conclusion}
We constructed EmoScene, a large-scale dual-space dataset that jointly represents categorical emotions, continuous VAD, measurable perceptual attributes, and contextual scene semantics. Scene-aware analyses reveal reproducible but non-deterministic associations between affective dimensions and perceptual attributes across data sources. As a reference implementation, AffectCtrl uses this supervision to enable categorical emotion control and continuous control over VAD, brightness, and saturation by learning conditioning-space residuals over frozen diffusion generators. Future work will extend controllable perceptual factors beyond brightness and saturation, broaden contextual coverage, and explore affect-aware editing and interactive generation.
\bibliography{main}

\clearpage
\setcounter{section}{0}
\setcounter{subsection}{0}
\setcounter{subsubsection}{0}
\setcounter{figure}{0}
\setcounter{table}{0}
\setcounter{equation}{0}
\renewcommand{\thesection}{S\arabic{section}}
\renewcommand{\thesubsection}{\thesection.\arabic{subsection}}
\renewcommand{\thesubsubsection}{\thesubsection.\arabic{subsubsection}}
\renewcommand{\thefigure}{S\arabic{figure}}
\renewcommand{\thetable}{S\arabic{table}}
\renewcommand{\theequation}{S\arabic{equation}}
\providecommand{\theHsection}{}
\providecommand{\theHsubsection}{}
\providecommand{\theHsubsubsection}{}
\providecommand{\theHfigure}{}
\providecommand{\theHtable}{}
\providecommand{\theHequation}{}
\renewcommand{\theHsection}{supp.\arabic{section}}
\renewcommand{\theHsubsection}{supp.\arabic{section}.\arabic{subsection}}
\renewcommand{\theHsubsubsection}{supp.\arabic{section}.\arabic{subsection}.\arabic{subsubsection}}
\renewcommand{\theHfigure}{supp.\arabic{figure}}
\renewcommand{\theHtable}{supp.\arabic{table}}
\renewcommand{\theHequation}{supp.\arabic{equation}}
\twocolumn[
\begin{@twocolumnfalse}
\begin{center}
{\LARGE\bfseries EmoScene: A Dual-Space Dataset for Controllable Affective Image Generation\par}
\vspace{0.6em}
{\Large\bfseries Supplementary Material\par}
\end{center}
\vspace{1em}
\end{@twocolumnfalse}
]
\raggedbottom

\section{Overview}
This supplement retains the construction, analysis, implementation, and evaluation details abbreviated in the seven-page main paper. Its organization follows the main-paper narrative so that each condensed claim has a corresponding reproducibility or evidence section:
Throughout the paper, \emph{dual-space} denotes two complementary representations of visual affect: an affective space comprising discrete emotions and continuous VAD, and a perceptual space comprising measurable appearance attributes. Contextual annotations ground both spaces in scene semantics and support stratified analysis rather than constituting a third representation space.
\begin{itemize}
  \item \textbf{Sec.~S2} expands dataset construction and release details, including the complete 11-step pipeline, annotation prompts, human verification, file formats, licensing, privacy, and opt-out procedures.
  \item \textbf{Sec.~S3} reports extended scene-aware analyses: VD/AD geometry, exact affect--perception estimates, adjustment and cross-source robustness, VAD-quartile shifts, and counterexamples to shallow visual heuristics.
  \item \textbf{Sec.~S4} specifies AffectCtrl training, including categorical token mapping, perceptual residual adaptation, continuous prompt residuals, losses, normalization, optimization, and inference.
  \item \textbf{Sec.~S5} provides the full experimental protocols and results compressed in the main paper: data sources, metrics, matched interfaces, target and cross-axis correlations, objective ablations, per-emotion results, human studies, and the interactive demo.
  \item \textbf{Sec.~S6} collects the qualitative evidence: matched VA comparison, eight-category generation, five-axis control trajectories, and extended EmoScene galleries with VAD scores, scene labels, and captions.
\end{itemize}

Code, configs, and EmoScene annotations will be released upon acceptance, together with pretrained AffectCtrl controllers and an inference demo (subject to licensing constraints).

\section{Dataset Construction Details}
\subsection{11-Step Annotation Pipeline}
\noindent\textbf{Image Collection.}
At Step~\GreenStep{1}, we conduct a cross-platform image collection process to build a large-scale multimodal emotional dataset with both diversity and high visual quality. The collection spans multiple visual domains, including photography, artistic creation, and social media. The data are sourced from professional photo-sharing platforms such as Unsplash, Pexels, Pixabay, and Flickr, as well as open archives from art institutions including the Metropolitan Museum of Art (Metmuseum). This multi-source strategy effectively integrates the natural authenticity of real-world scenes with the aesthetic richness of artistic works, ensuring broad coverage across themes, styles, and cultural contexts.
To improve the relevance of collected content, we adopt a hybrid strategy combining keyword-based retrieval with topic grouping, enabling precise selection of images closely related to affective experiences. The collection encompasses diverse categories such as natural landscapes, urban environments, human activity scenes, social interactions, and technological or artistic domains. In total, approximately 1.6 million images were gathered, covering 345 representative scene categories, providing a solid foundation for subsequent quality filtering and multi-dimensional annotation.

\noindent\textbf{Image Filtering.}
At Steps~\GreenStep{2}–\GreenStep{3}, we perform automated filtering and consistency validation to ensure both image quality and semantic accuracy of scene labels.
At Step~\GreenStep{2}, an aesthetic assessment model and a sharpness detection algorithm are employed to compute a unified quality score for each image. Samples with blur, exposure imbalance, or low resolution are automatically removed, improving overall visual quality and dataset stability.
At Step~\GreenStep{3}, scene-label consistency is verified using a ResNet-50 model pretrained on Places365. Images whose predicted and original labels align are retained or updated, while inconsistent samples are discarded. We further employ a CLIP-based image–text similarity check to eliminate low-matching pairs, enhancing the semantic alignment between visual content and scene annotations.
The resulting dataset exhibits substantially improved clarity, scene accuracy, and semantic coherence, providing high-quality input for the subsequent dual-space annotation stage.

\noindent\textbf{Entity Recognition and Human Analysis.}
At Steps~\GreenStep{4}–\GreenStep{5}, we perform fine-grained annotation from the perspective of entity understanding to establish the semantic and contextual foundation required for affective image generation.
At Step~\GreenStep{4}, all human subjects in the image are detected, and their attributes—including gender, age, and dominant facial expression—are identified. We employ YOLOv8n for human detection and use facial analysis models to estimate gender and age following the JRDB-Social age taxonomy\cite{jahangard2024jrdb}, while also classifying primary emotional expressions. To further capture interaction dynamics, InternVL3-8B is integrated with a Chain-of-Thought (CoT) reasoning strategy to infer human–scene relationships (e.g., “comforting,” “gazing,” “avoiding”). This step enables the transition from static appearance to dynamic semantics, enriching the behavioral logic and narrative coherence in emotion-driven image generation.
At Step~\GreenStep{5}, a YOLO model pretrained on Object365 is utilized to identify non-human objects and record their categories and counts. This process provides contextual environmental information that complements human-centered semantics, forming a scene-aware layer that supplies structured input for subsequent text generation.

\noindent\textbf{Stylized Description Generation.} 
At Step~\GreenStep{6}, we model the language layer to generate diverse textual descriptions that strengthen the emotional correspondence between visual semantics and linguistic expression. Specifically, the Human-Aware Modeling (HAM) framework and a multimodal large language model (MLLM) are jointly employed: HAM produces concise short descriptions to capture core semantics and emotional cues, while the MLLM integrates human–object interaction reasoning to generate fine-grained long descriptions, enriching both narrative coherence and emotional depth.
To further enhance linguistic expressiveness, we extract and cluster distinct language styles from the MSCOCO corpus, learn style prompt vectors, and fine-tune the MLLM to generate text conditioned on specific stylistic cues. For each image, the model produces five randomly sampled descriptions across different styles, forming a one-to-many linguistic mapping. This design maintains semantic consistency while improving controllability over emotional semantics and increasing diversity in language generation.

\noindent\textbf{Affective Space Annotation.}
At Steps~\GreenStep{7}–\GreenStep{8}, each image is annotated with both discrete and continuous emotional representations, forming a comprehensive affective space.
At Step~\GreenStep{7}, we perform eight-class discrete emotion classification to assign each image a clear affective label. Qwen2.5 and InternVL3-8B vision–language models are used together with the CoT reasoning strategy, enabling the models to infer emotion from image semantics, human states, and overall scene atmosphere. The model outputs one of eight core emotions, or “Neutral” when the emotion is weak, ensuring that each image conveys a distinct and perceivable affective intent.
At Step~\GreenStep{8}, we construct continuous emotional representations in the VAD space to capture fine-grained intensity variations. The same Qwen2.5 and InternVL3-8B models are applied, with customized CoT prompts for each dimension to reason about scene content and human interactions before assigning scores. For the more abstract Dominance dimension, CoT reasoning significantly improves consistency across predictions. We adopt a human-aligned annotation protocol in which models assign scores on a discrete [1, 9] scale following the Self-Assessment Manikin (SAM) framework. The final VAD scores are obtained by taking the weighted average of both model outputs, reducing single-model bias and improving overall reliability.

\noindent\textbf{Perceptual Space Annotation.}
At Steps~\GreenStep{9}–\GreenStep{10}, we extract low-level perceptual features from color and structural dimensions to capture visual cues that strongly influence affective perception.
At Step~\GreenStep{9}, we annotate the color attributes of each image.
To balance perceptual accuracy and feature richness, we adopt a multi-scale color analysis strategy.Each image is first converted into the perceptually uniform CIE Lab space, and the Euclidean distance between each pixel and 11 reference colors is computed to quantify the dominant hues and overall color distribution.We then calculate global HSV statistics to describe the color atmosphere of the image, where saturation and brightness (value) are later analyzed for their associations with VAD. This supports a measurable, non-deterministic characterization of how color composition co-varies with affective dimensions.
At Step~\GreenStep{10}, we quantify structural attributes that modulate emotional perception. Canny edge detection is applied to extract image contours, and the mean angular deviation between edge vectors is computed to encode the perceived “softness” or “rigidity” of the scene as a continuous measure. In addition, visual complexity is assessed by integrating two complementary metrics: information entropy (texture complexity) and edge density (structural complexity). These jointly characterize the relationship between scene intricacy—ranging from tranquil to cluttered—and the arousal dimension. Together, the color and structural features form the perceptual space, providing essential support for dual-space affective modeling.

\begin{figure*}[!t]
\centering
\begin{minipage}{1.0\textwidth}
\small
\begin{verbatim}
Your task is to act as an emotional analysis expert and provide a comprehensive 
assessment of the provided image, analyzing its conveyed three emotional 
dimensions (Valence, Arousal, Dominance) and the main discrete emotion category.

Please follow these steps to complete the analysis:

Part One: Scene and Subject Analysis (Text Output Only)
1. Scene Description: Briefly describe the main content and scenario in the image.
2. Subject Analysis: Analyze the state, posture, expression, and interaction with 
  the environment of the main subject(s) (person, animal, or object). Determine 
  whether it appears active or passive, powerful or weak.

Part Two: Three Dimensions and Discrete Emotion Assessment (Text and Final Scores Only)
You need to synthesize all visual cues in the image and deduce the three 
dimensional indices and one discrete emotion category.

1. Valence (Pleasure): 1=Extremely Negative, 5=Neutral, 9=Extremely Pleasant
2. Arousal (Activation): 1=Extremely Calm, 5=Normal, 9=Extremely Exciting
3. Dominance (Control): 1=Completely Powerless, 5=Neutral, 9=Complete Control
4. Primary Emotion: amusement, anger, awe, contentment, disgust, excitement,
  fear, sadness. Use neutral if no significant emotion is present. 
  Use unknown if it cannot be determined.

Scoring Rationale (VAD): Based on your analysis, explain why you gave these 
three specific VAD scores.

Final Output Format Requirement:
Please strictly output your analysis results in the following format.

Valence score: [1-9 integer]
Arousal score: [1-9 integer]
Dominance score: [1-9 integer]
Primary emotion: [amusement, anger, awe, contentment, disgust, excitement, fear, 
sadness, neutral, or unknown]

Analysis Process:
(Output your Scene Description, Subject Analysis, and Scoring Rationale here)
\end{verbatim}
\end{minipage}
\caption{Prompt used for CoT-style VAD and discrete emotion annotation.}
\label{fig:prompt}
\end{figure*}

\noindent\textbf{Human-in-the-Loop Quality Control.}
At Step~\GreenStep{11}, we establish a quality control mechanism that integrates human verification with active learning to ensure the reliability and consistency of dataset annotations.
The verification process is conducted by trained human annotators who evaluate model predictions and correct potential biases, forming a high-quality feedback loop between automated annotation and human judgment.
For discrete emotion classification, when the two models (Qwen2.5 and InternVL3-8B) produce consistent results, annotators confirm whether the predicted emotion aligns with their own perception; if disagreement arises, they provide an independent judgment with justification. In cases where the two models disagree, annotators either support one of the predictions or reassign a new label, documenting the rationale to enhance consistency and transparency.
For continuous emotion annotation (VAD), annotators comprehensively assess image color, composition, and semantic content to verify the averaged VAD values produced by the models. If a notable deviation from perceived affective intensity is identified, manual adjustments are made along with clear criteria and reasoning to ensure that the corrected values faithfully represent the emotional strength and polarity of the image.
Finally, all human-corrected results are fed back into the automatic annotation system, forming an iterative active learning cycle that maintains EmoScene’s high semantic, perceptual, and affective consistency and credibility.

\noindent\textbf{Representative dual-space examples.}
The representative pair in the main paper illustrates how these fields are read jointly. The amusement-park scene combines high valence and arousal with greater brightness and complexity and a lively semantic context; the mist-forest scene combines lower valence with a darker, lower-complexity profile and gloomy context. These examples illustrate aligned annotations rather than deterministic visual rules.

\subsection{Prompt Templates and CoT Rules for VAD and Discrete Emotions}
To obtain consistent, fine-grained affect annotations beyond coarse labels, we adopt a chain-of-thought style assessment prompt (detailed in Figure $\ref{fig:prompt}$). This prompt asks an annotator model to first describe the scene and subjects, then reason about the affective cues present, and finally output the discrete emotion and three integer scores for Valence, Arousal, and Dominance (VAD) on a 1–9 scale. We use a single, fixed template for all images. We parse the four \emph{final lines} with regex to obtain the three integer scores and the primary emotion (from a fixed vocabulary plus neutral and unknown). Non-integer or out-of-range values are clamped to ([1-9]) and re-queried once.

\subsection{Quality Control, Conflict Resolution, and Human-in-the-loop}
\label{sec:qc-hitl}

\begin{figure}[!t] 
  \centering
  \includegraphics[width=\linewidth]{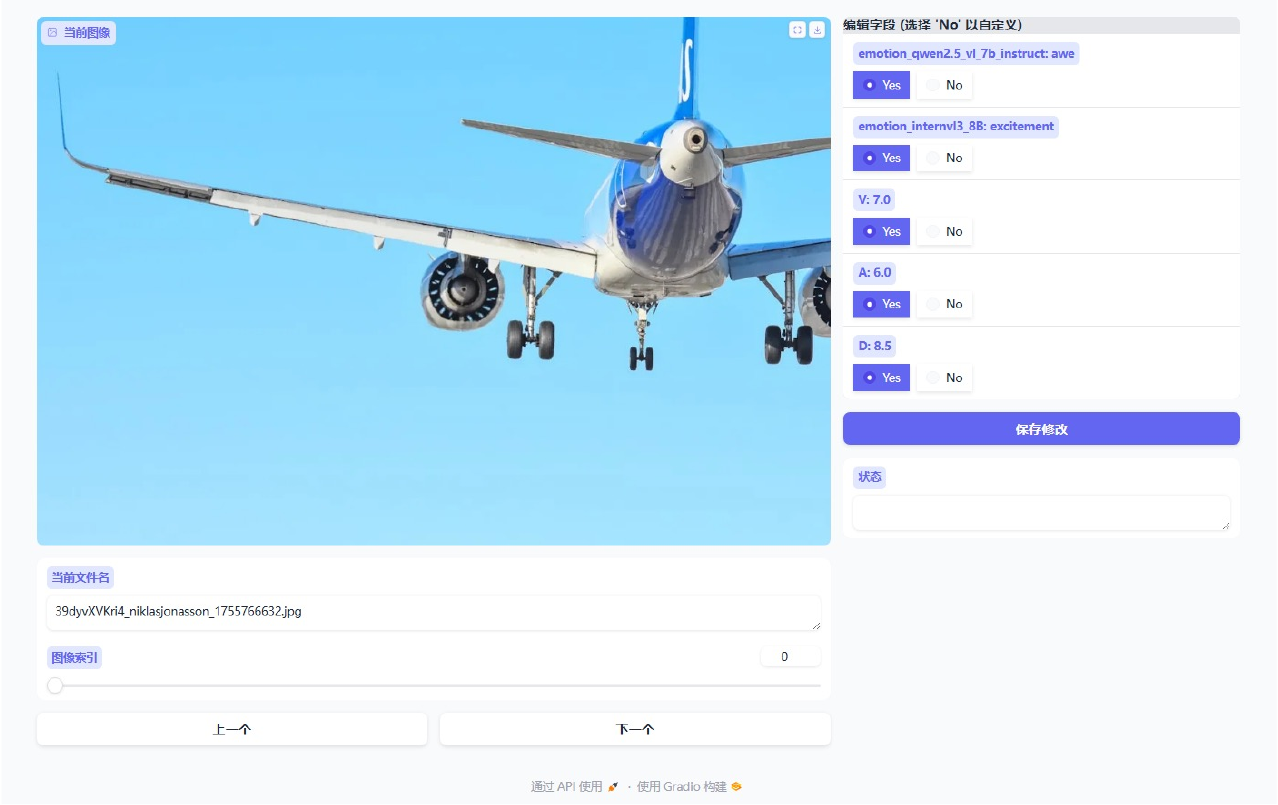}
  \caption{\textbf{Human-in-the-loop verification UI.}
  Left: the current image. Right: machine annotations to be verified:
  two discrete-emotion suggestions (e.g., from different annotators/models) and
  three V/A/D scores. The reviewer only selects \emph{Yes/No} for each field.
  Selecting \emph{No} requires a brief rationale in the comment box and moves
  the sample to the re-check queue; \emph{Yes} accepts the value as final.}
  \label{fig:hitl}
\end{figure}

\noindent\textbf{Protocol.}
In addition to this large-scale binary audit, we conduct an independent multi-rater evaluation on 3,000 images sampled through source$\times$scene$\times$emotion stratification. Each image is independently annotated by three human annotators, providing the repeated judgments used to assess inter-rater agreement.
For the large-scale audit, we implement a rigorous binary verification mechanism for both discrete emotions and continuous VAD scores, as shown in Figure~\ref{fig:hitl}:
\begin{itemize}
  \item \textbf{Validation Logic.} Reviewers accept (\emph{Yes}) or reject (\emph{No}) each annotation field independently. A rejection triggers a mandatory text input for the rationale (e.g., ``arousal inconsistent with calm scene''), ensuring that negative feedback is actionable.
  \item \textbf{Refinement Loop.} Validated entries are finalized immediately. Rejected items are flagged and queued for a second pass, involving either re-annotation with different seeds or manual correction by senior experts, until they pass verification.
  \item \textbf{Conflict Resolution.} When presented with competing discrete emotion candidates, reviewers validate the most appropriate label and reject the alternative with a brief justification. VAD scores are assessed strictly against visual evidence.
  \item \textbf{Audit Trail.} We log all decisions, timestamps, and rejection rationales. These records enable continuous auditing and prompt refinement.
\end{itemize}

\noindent\textbf{Outcome.}
Samples with any No remain pending until corrected and re-verified.
Samples with all Yes are accepted as final. This process provides a
transparent trail from machine suggestion to human approval and ensures that
both discrete emotion and VAD labels are explicitly justified when corrected,
which improves reliability for downstream analyses and training.

\subsection{Field Definitions and File Formats}
\noindent\textbf{Directory layout and pairing.}
The dataset is organized by scene. For each image, the JPEG file and its annotation JSON share the same \texttt{<stem>} and reside in the same scene folder:
\begin{verbatim}
EmoScene/
  beach/
    <stem>.jpg    # image file
    <stem>.json   # annotation file
  forest/
    <stem>.jpg
    <stem>.json
  street/
    ...
\end{verbatim}
EmoScene does not prescribe a single official split; the task-specific data
construction used in our experiments is documented in
Sec.~\ref{sec:supp_data_sources}.

\begin{figure*}[!t]
  \centering
  \begin{minipage}[t]{0.48\linewidth}
  \centering
    \vspace{0pt} 
    \includegraphics[width=\linewidth]{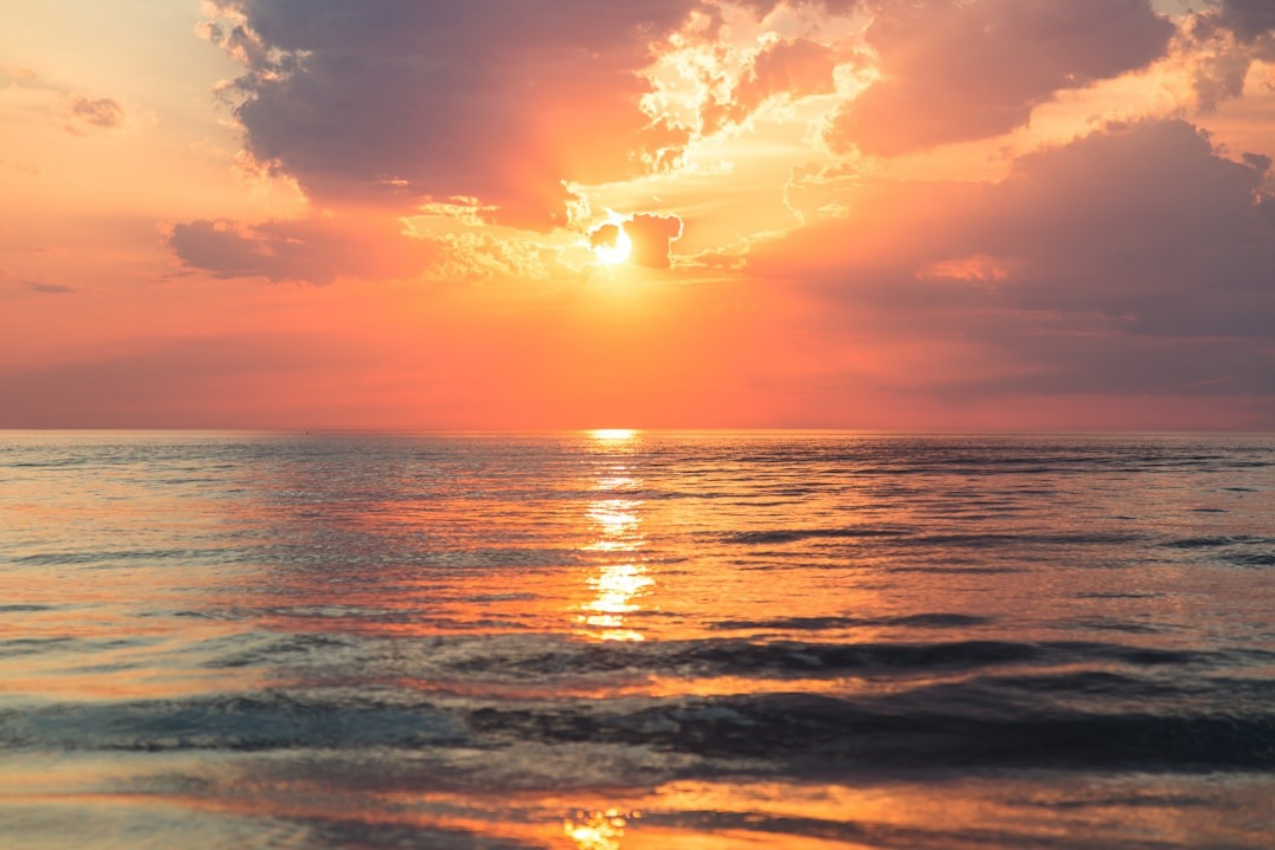}
  \end{minipage}
  \hfill
  \begin{minipage}[t]{0.48\linewidth}
    \centering
    \scriptsize 
    \begin{verbatim}
{
  "clip_similarity": 0.2081,
  "scene": "beach",
  "emotion": "awe",
  "description_InternVL3-8B":
    "The image captures a serene sunset ...",
  "color_proportion": {
    "Black": 0.06, "White": 0.03, "Orange": 0.12,
    "Pink": 0.17, "Brown": 0.18, "Gray": 0.44,
    "Red": 0.0, "Green": 0.0, "Blue": 0.0,
    "Yellow": 0.0, "Purple": 0.0
  },
  "average_color": {
    "hue": 20, "saturation": 45, "value": 69},
  "people_count": 0, "persons": [],
  "objects": {},
  "curvilinearity": 0.0986,
  "complexity_entropy": 0.9404,
  "complexity_edge_density": 0.0392,
  "aesthetic_score": 6.4474, "liqe_score": 3.3281,
  "valence_internvl3_8B": 8, 
  "arousal_internvl3_8B": 7,
  "emotion_internvl3_8B": "awe",
  "valence_qwen2.5_vl_7b_instruct": 8,
  "arousal_qwen2.5_vl_7b_instruct": 7,
  "dominance_qwen2.5_vl_7b_instruct": 7,
  "emotion_qwen2.5_vl_7b_instruct": "awe"
}
\end{verbatim}
  \end{minipage}  
  \caption{\textbf{Pairing example.} In the \texttt{beach} folder, the image
  (\texttt{.jpg}) and its annotation (\texttt{.json}) share the same \texttt{<stem>};
  the bottom block shows a truncated JSON corresponding to the image.}
  \label{fig:example}
\end{figure*}

\noindent\textbf{Pairing illustration.}
To make the directory layout and one-to-one pairing concrete, Fig.~\ref{fig:example}
shows a sample from the \texttt{beach} scene folder: the JPEG and its annotation
JSON share the same \texttt{<stem>} and reside in the same directory.
The accompanying JSON snippet (bottom) is representative of the fields released
with the dataset; per-model affect predictions (keys with model suffixes) are
kept verbatim for transparency, alongside aggregated labels where applicable.

\subsection{Data Licensing, Privacy, and Opt-out}
\label{sec:supp_licensing}
\noindent\textbf{Data sources and licensing metadata.}
EmoScene is collected from multiple public platforms and open archives (e.g., Unsplash, Pexels, Pixabay, Flickr, and The Met Open Access) to cover both real-world scenes and artistic imagery.
Following the release policy, we distribute the dataset in an annotation-first manner with paired JSON metadata, including (i) source platform and (ii) original URL/identifier when available. We only release subsets for which the associated usage terms can be reasonably established, and include licensing/usage references when provided by the host platform.
We will distribute the dataset in an \emph{annotation-first} manner (image file + paired JSON metadata) and only release subsets that satisfy the corresponding licensing constraints.

\noindent\textbf{Human subjects and privacy.}
EmoScene contains diverse scenes and may include images with people.
To support transparency, we store a coarse 'people\_count' field and related non-identifying contextual attributes in the paired JSON metadata.
We do not provide identity labels, biometric identifiers, or any face recognition annotations, and the dataset is not intended for re-identification, surveillance, or inferring sensitive personal attributes.

\noindent\textbf{De-duplication and near-duplicate removal.}
To reduce privacy risks and improve benchmark integrity, we apply de-duplication before release.
We remove exact duplicates and near-duplicates using a combination of perceptual hashing and embedding-based similarity (e.g., CLIP image embeddings), and we will provide a duplicate-list (or hash list) in the release package for transparency.

\noindent\textbf{Opt-out mechanism.}
We will provide an opt-out request channel at release time.
A requester can submit an image URL/identifier (or the image itself), and we will match it against our metadata/hashes to remove the corresponding sample(s) from future releases and subsequent dataset versions.
We will also publish instructions and response-time expectations in the dataset usage policy.

\noindent\textbf{Potential misuse and intended use.}
EmoScene is released to facilitate research on affect--perception interactions, affective understanding, and controllable generation.
We caution that affect labels are inherently subjective and can reflect cultural or contextual biases.
We explicitly discourage using the dataset to generate manipulative or deceptive content, and we will include a usage policy that prohibits applications involving targeted persuasion or harassment.
\section{Extended Scene-Aware Dataset Analyses}
This section expands the main-paper analysis supporting Contribution~2: a large-scale, scene-aware characterization of how categorical emotion, continuous VAD, and measurable perceptual attributes co-vary in natural imagery. All results below use the same frozen cohorts as main-paper Figs.~4--5. Samples are selected by a fixed hash before statistics are computed; no image or seed is selected according to an observed effect.

\subsection{Affective Geometry Beyond the VA Plane}
The main paper summarizes class structure in the Valence--Arousal plane. Figures~\ref{fig:supp-vd} and~\ref{fig:supp-ad} apply the same class-normalized centroid and covariance-ellipse visualization to Valence--Dominance and Arousal--Dominance. Positive classes cluster at higher valence and moderately higher dominance, whereas fear, sadness, and disgust occupy lower-dominance regions. The AD view further separates high-arousal anger and excitement from low-dominance fear and disgust. Ellipse overlap remains substantial, which is expected because the discrete categories partition a continuous affective space.

\subsection{Frozen Cohorts and Statistical Protocol}
Table~\ref{tab:supp_analysis_cohorts} records the two frozen analysis cohorts. The profile cohort caps each emotion at 2,500 and allocates samples proportionally across source$\times$scene strata. The association cohort caps each eligible source$\times$scene cell at 20, preventing large scene/source cells from dominating continuous correlations. The resulting association cohort is nearly balanced by source: Flickr/Pexels/Pixabay/Unsplash contribute 6,766/6,861/6,750/6,875 images.

\begin{table*}[!t]
\centering
\small
\setlength{\tabcolsep}{4pt}
\begin{tabular}{lrrrl}
\toprule
Analysis cohort & Images & Scenes & Source$\times$scene cells & Selection role \\
\midrule
Profile cohort & 18,085 & 342 & 1,292 & At most 2,500 per emotion; proportional stratified hash sample \\
Association cohort & 27,252 & 344 & 1,374 & At most 20 per eligible source$\times$scene cell \\
\bottomrule
\end{tabular}
\caption{Frozen cohorts used by the scene-aware dataset analysis. Both cohorts are selected a priori by fixed hash from records with complete, protocol-compatible affective and perceptual annotations. Source$\times$scene cells with fewer than five records are excluded from the association cohort.}
\label{tab:supp_analysis_cohorts}
\end{table*}

For category profiles, each perceptual feature is globally standardized and then centered by its matched source$\times$scene mean. For continuous associations, both the perceptual feature and VAD target are centered within the same source$\times$scene cell before computing Pearson correlation. We use 2,000 scene-cluster bootstrap resamples for 95\% intervals and two-sided empirical zero-tail tests. Benjamini--Hochberg correction is applied separately to the 54 profile cells and 18 association cells; asterisks require both an interval excluding zero and adjusted $q<.05$.

\subsection{Complete Affect--Perception Estimates}
The main paper visualizes the scene-aware category profiles and continuous associations. Tables~\ref{tab:supp_profile_color}--\ref{tab:supp_association_exact} provide the corresponding unrounded values, with each cell reporting an estimate $[95\%\ \mathrm{CI}]$ followed by BH-adjusted $q$. These tables expose estimate uncertainty and make the main-paper visualization fully auditable.

\begin{table*}[!t]
\centering
\scriptsize
\setlength{\tabcolsep}{3pt}
\begin{tabular}{lccc}
\toprule
Emotion & Brightness & Saturation & Hue warmth \\
\midrule
Contentment & $+.125\,[+.093,+.159]$; .002 & $+.152\,[+.116,+.185]$; .002 & $-.003\,[-.034,+.029]$; .883 \\
Amusement & $+.215\,[+.159,+.286]$; .002 & $+.120\,[+.067,+.169]$; .004 & $+.029\,[+.002,+.064]$; .060 \\
Excitement & $-.007\,[-.051,+.042]$; .878 & $+.168\,[+.108,+.246]$; .002 & $+.048\,[+.004,+.094]$; .057 \\
Awe & $0.000\,[-.045,+.046]$; .993 & $+.253\,[+.199,+.309]$; .002 & $-.064\,[-.108,-.018]$; .017 \\
Anger & $-.136\,[-.500,-.060]$; .019 & $-.135\,[-.364,+.121]$; .206 & $-.059\,[-.447,+.033]$; .183 \\
Fear & $-.796\,[-.971,-.631]$; .002 & $-.004\,[-.120,+.112]$; .973 & $-.270\,[-.398,-.148]$; .002 \\
Sadness & $-.358\,[-.444,-.285]$; .002 & $-.280\,[-.349,-.222]$; .002 & $+.036\,[-.012,+.082]$; .183 \\
Disgust & $+.037\,[-.154,+.080]$; .683 & $+.068\,[-.035,+.184]$; .155 & $+.034\,[-.005,+.178]$; .115 \\
Neutral & $-.070\,[-.110,-.036]$; .002 & $-.128\,[-.167,-.091]$; .002 & $+.012\,[-.031,+.052]$; .668 \\
\bottomrule
\end{tabular}
\caption{Exact scene-adjusted category profiles for the three color-related perceptual features. Entries are residual z-score $[95\%\ \mathrm{CI}]$; BH-adjusted $q$.}
\label{tab:supp_profile_color}
\end{table*}

\begin{table*}[!t]
\centering
\scriptsize
\setlength{\tabcolsep}{3pt}
\begin{tabular}{lccc}
\toprule
Emotion & Texture entropy & Edge density & Curvilinearity \\
\midrule
Contentment & $+.153\,[+.126,+.181]$; .002 & $+.087\,[+.058,+.118]$; .002 & $+.108\,[+.077,+.141]$; .002 \\
Amusement & $+.186\,[+.113,+.271]$; .002 & $+.137\,[+.083,+.189]$; .002 & $+.054\,[+.021,+.093]$; .004 \\
Excitement & $-.032\,[-.097,+.024]$; .314 & $+.086\,[+.043,+.132]$; .002 & $+.108\,[+.051,+.175]$; .002 \\
Awe & $+.061\,[+.017,+.100]$; .020 & $+.016\,[-.031,+.066]$; .591 & $+.038\,[-.006,+.076]$; .129 \\
Anger & $-.045\,[-.358,+.287]$; .626 & $-.115\,[-.179,+.065]$; .198 & $+.074\,[-.112,+.112]$; .470 \\
Fear & $-.941\,[-1.229,-.671]$; .002 & $-.417\,[-.654,-.216]$; .002 & $-.009\,[-.150,+.085]$; .883 \\
Sadness & $-.127\,[-.196,-.072]$; .002 & $-.093\,[-.145,-.042]$; .002 & $+.056\,[+.006,+.110]$; .048 \\
Disgust & $+.156\,[+.111,+.327]$; .005 & $+.456\,[+.373,+.583]$; .002 & $+.365\,[+.302,+.649]$; .002 \\
Neutral & $-.079\,[-.118,-.045]$; .002 & $-.056\,[-.090,-.023]$; .005 & $-.089\,[-.123,-.059]$; .002 \\
\bottomrule
\end{tabular}
\caption{Exact scene-adjusted category profiles for the three structural perceptual features. Entries are residual z-score $[95\%\ \mathrm{CI}]$; BH-adjusted $q$.}
\label{tab:supp_profile_structure}
\medskip

\begin{tabular}{lccc}
\toprule
Feature & Valence & Arousal & Dominance \\
\midrule
Brightness & $+.177\,[+.159,+.194]$; .001 & $+.013\,[-.007,+.033]$; .230 & $+.118\,[+.103,+.135]$; .001 \\
Saturation & $+.223\,[+.200,+.244]$; .001 & $+.177\,[+.158,+.198]$; .001 & $+.147\,[+.130,+.165]$; .001 \\
Hue warmth & $-.023\,[-.041,-.005]$; .012 & $-.008\,[-.026,+.009]$; .375 & $-.029\,[-.046,-.012]$; .002 \\
Texture entropy & $+.132\,[+.115,+.149]$; .001 & $+.059\,[+.038,+.079]$; .001 & $+.096\,[+.077,+.114]$; .001 \\
Edge density & $+.110\,[+.089,+.131]$; .001 & $+.136\,[+.119,+.155]$; .001 & $+.098\,[+.074,+.119]$; .001 \\
Curvilinearity & $+.110\,[+.091,+.130]$; .001 & $+.146\,[+.127,+.165]$; .001 & $+.071\,[+.053,+.089]$; .001 \\
\bottomrule
\end{tabular}
\caption{Exact source$\times$scene-adjusted continuous associations. Entries are Pearson $r$ $[95\%\ \mathrm{CI}]$; BH-adjusted $q$.}
\label{tab:supp_association_exact}
\medskip

\begin{minipage}[t]{0.48\textwidth}
  \centering
  \includegraphics[width=\linewidth]{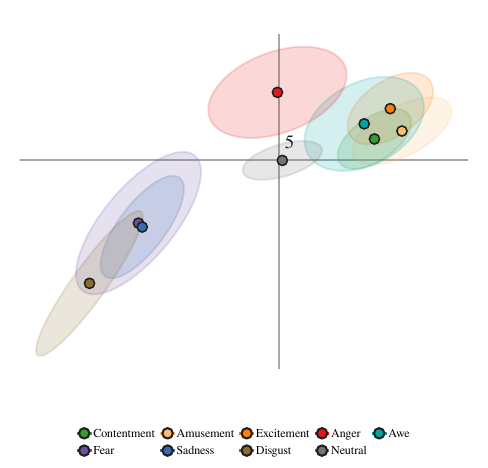}
  \captionof{figure}{\textbf{Class-normalized Valence--Dominance structure.} Markers are per-emotion centroids and ellipses are the 50\% covariance regions in the frozen, source--scene-stratified profile cohort. Colors and visual encoding match main-paper Fig.~4; ellipse size reflects within-class spread rather than class frequency.}
  \label{fig:supp-vd}
\end{minipage}\hfill
\begin{minipage}[t]{0.48\textwidth}
  \centering
  \includegraphics[width=\linewidth]{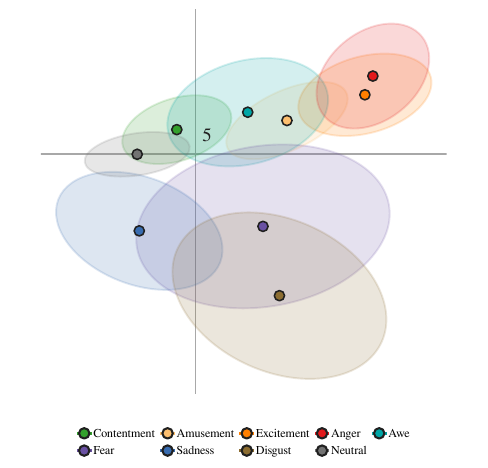}
  \captionof{figure}{\textbf{Class-normalized Arousal--Dominance structure.} The plotting protocol and frozen cohort are identical to Fig.~\ref{fig:supp-vd}, with arousal replacing valence on the horizontal axis. The view exposes variation along dominance that is not visible in the main-paper VA projection.}
  \label{fig:supp-ad}
\end{minipage}
\end{table*}

\subsection{Adjustment and Cross-Source Robustness}
\begin{figure*}[!t]
  \centering
  \includegraphics[width=0.96\textwidth]{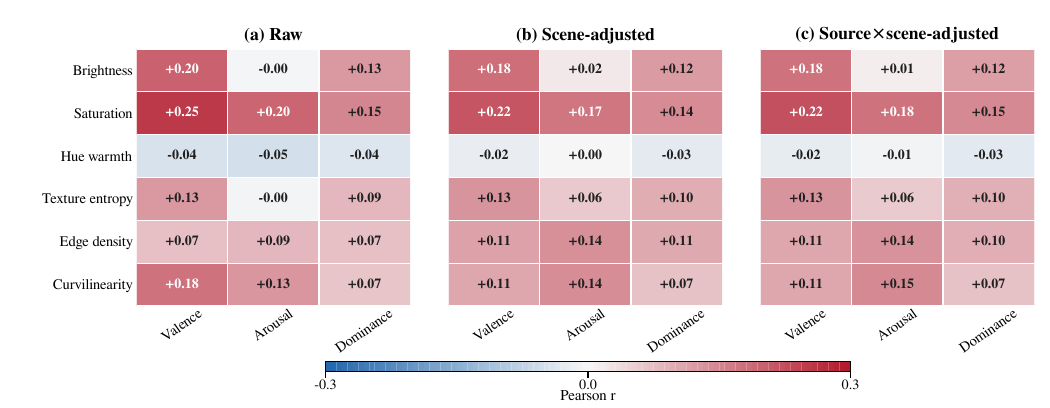}
  \caption{\textbf{Robustness to source and scene composition.} Pearson correlations are computed on the same frozen 27,252-image association cohort. (a) Raw correlations; (b) correlations after centering both variables by scene; (c) correlations after centering within each source$\times$scene cell. Panel (c) reproduces the estimates in main-paper Fig.~6(b) to provide a direct comparison across adjustment protocols. Stable signs and similar magnitudes indicate that the reported coupling is not an artifact of a few dominant sources or scene categories.}
  \label{fig:supp_adjustment_robustness}
\end{figure*}

Figure~\ref{fig:supp_adjustment_robustness} compares raw correlations with scene-only and source$\times$scene adjustment. The principal relationships survive both controls: saturation--valence changes from $.254$ to $.223$, saturation--arousal from $.196$ to $.177$, and brightness--valence from $.201$ to $.177$. Curvilinearity--arousal slightly increases from $.131$ to $.146$, while the already small hue-warmth correlations move closer to zero. Thus, source and scene composition affect magnitudes but do not create the main affect--perception pattern.

\begin{figure*}[!t]
  \centering
  \includegraphics[width=0.96\textwidth]{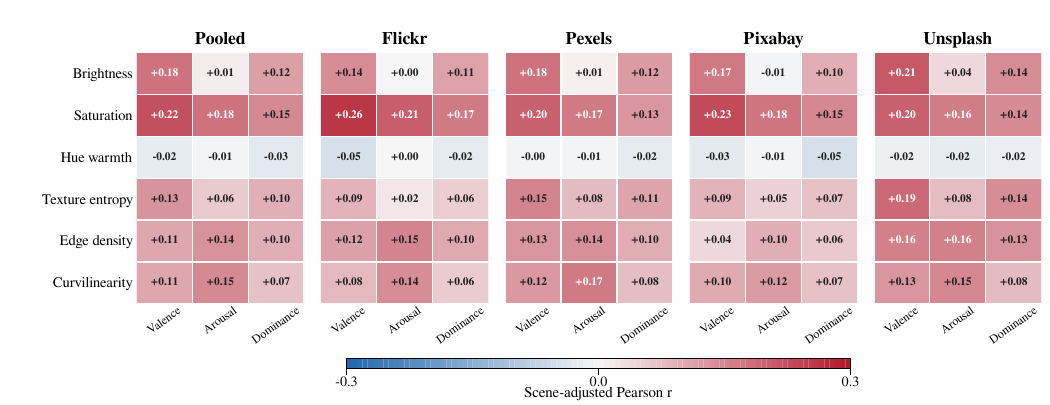}
  \caption{\textbf{Cross-source robustness of scene-adjusted associations.} The pooled source$\times$scene-adjusted correlations are followed by estimates computed independently within Flickr, Pexels, Pixabay, and Unsplash after scene centering. The four source panels are diagnostic replications rather than additional significance tests; formal scene-cluster intervals for the pooled analysis are reported in Table~\ref{tab:supp_association_exact}.}
  \label{fig:supp_source_robustness}
\end{figure*}

Figure~\ref{fig:supp_source_robustness} repeats the source$\times$scene-adjusted analysis separately for each collection source. Saturation remains positively associated with valence ($r=.20$--$.26$) and arousal ($r=.16$--$.21$) in every source; brightness--valence remains positive ($r=.14$--$.21$); and edge density and curvilinearity consistently track arousal. Source-specific effect sizes differ, as expected from photographic style and scene coverage, but the central directions reproduce across all four platforms.

\subsection{A Perceptual Bridge Along Continuous VAD}
\begin{figure*}[!t]
  \centering
  \includegraphics[width=0.90\textwidth]{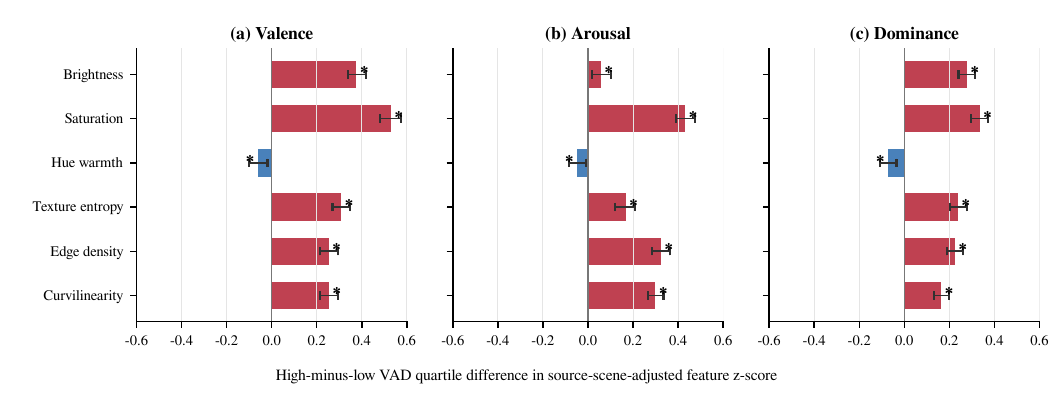}
  \caption{\textbf{Perceptual shifts between high and low VAD quartiles.} The three panels show valence, arousal, and dominance from left to right. Bars report the upper-minus-lower VAD-quartile difference in source$\times$scene-adjusted perceptual feature z-score; error bars are 95\% scene-cluster bootstrap intervals from 2,000 resamples. Asterisks require an interval excluding zero and BH-adjusted $q<.05$ across all 18 comparisons. This difference-in-means view complements the continuous correlations summarized in the main paper.}
  \label{fig:supp_vad_bridge}
\end{figure*}

Correlations summarize monotonic co-variation but do not show the size of the appearance shift between affective extremes. Figure~\ref{fig:supp_vad_bridge} therefore compares the upper and lower quartiles of each source$\times$scene-adjusted VAD dimension. High-valence images are especially more saturated ($+.529z$), brighter ($+.376z$), and more textured ($+.309z$). High arousal has its largest shifts in saturation ($+.432z$), edge density ($+.324z$), and curvilinearity ($+.299z$). Dominance also has distributed perceptual associations, led by saturation ($+.334z$) and brightness ($+.278z$), rather than a single exclusive low-level carrier.

\subsection{Counterexamples to Shallow Visual Heuristics}
\begin{figure*}[!t]
  \centering
  \includegraphics[width=0.96\textwidth]{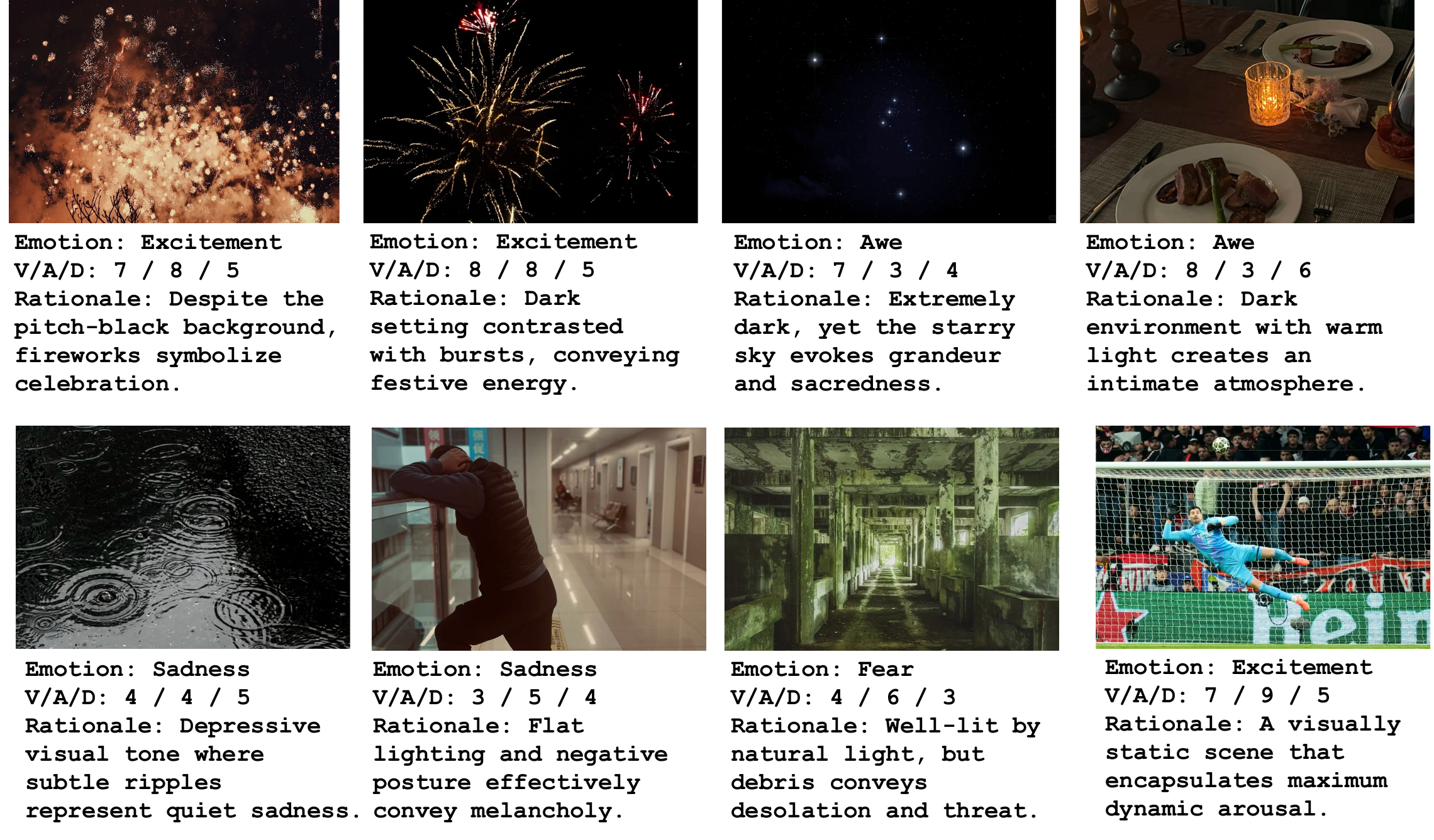}
  \caption{\textbf{Counterexamples to shallow visual heuristics.} Eight challenging samples pair VAD scores with annotation rationales. The top row contains low-luminance scenes with positive or high-arousal affect; the bottom row contains bright or visually static scenes whose semantics convey negative affect or high arousal. The gallery illustrates that the scene-aware statistical associations are non-deterministic and should not be interpreted as rules such as ``low luminance implies negative valence.''}
  \label{fig:bias}
\end{figure*}

The population-level associations above are statistical tendencies, not deterministic rules. Figure~\ref{fig:bias} shows counterexamples to shortcuts such as ``dark means negative'' or ``bright means positive'': fireworks, a night sky, and a candle-lit dinner receive positive affect labels despite low luminance, whereas bright or visually static scenes can convey sadness, fear, or high arousal through their semantics. These examples do not establish a causal annotation mechanism, but they demonstrate why both perceptual and contextual annotations are needed to characterize affect in diverse scenes.

\clearpage
\twocolumn


\section{AffectCtrl Training Details}
\label{sec:supp_training}

AffectCtrl keeps the image generator frozen and learns only small controllers in the conditioning space.  This section provides the implementation-level view that complements Sec.~4 of the main paper.  We use two interface-aligned branches: a categorical branch for EmoGen-style emotion-token control, and a continuous branch for SDXL prompt residual control.

\subsection{Categorical Branch}
\label{sec:supp_training_categorical}
\noindent\textbf{EmoScene-compatible EmoGen training.}
We first convert EmoScene into an EmoGen-compatible training split.  Each training sample contains an emotion label, a scene/object-balanced prompt, and the corresponding image.  The generator, EmoSpace representation, and auxiliary frozen networks used by the EmoGen-style objective are kept unchanged.  Given an emotion label $y$, we retrieve either the EmoSpace prototype or a sampled emotion embedding $z_e$ and train an emotion mapper
\begin{equation}
  t_{\mathrm{aff}} = M_{\mathrm{cat}}(z_e),
\end{equation}
where $t_{\mathrm{aff}}$ is the affective token inserted into the frozen generator condition.  This stage isolates the contribution of EmoScene categorical supervision under the same interface as EmoGen.

\noindent\textbf{Perceptual residual adapter.}
After training $M_{\mathrm{cat}}$, we freeze it and train a residual adapter that conditions on both the emotion embedding and perceptual attributes.  We use continuous mean HSV brightness and saturation values as the perceptual control vector $p=[b,s]$.  The adapter predicts
\begin{equation}
  r_{\mathrm{per}} = M_{\mathrm{per}}(z_e, p), \qquad
  t_{\mathrm{final}} = t_{\mathrm{aff}} + \lambda r_{\mathrm{per}}.
\end{equation}
The residual is deliberately additive: the categorical token preserves the target emotion direction, while the perceptual residual changes how the emotion is visually expressed.  During inference, $p$ can be specified explicitly or sampled from the target emotion's EmoScene perceptual distribution.

\noindent\textbf{Architecture details.}
The categorical mapper $M_{\mathrm{cat}}$ is a two-layer MLP with dimensions $768\!\rightarrow\!1024\!\rightarrow\!768$ and a ReLU activation between the two linear layers.  The perceptual residual adapter first maps the two-dimensional brightness--saturation vector through a $2\!\rightarrow\!128\!\rightarrow\!128$ encoder.  The resulting feature is concatenated with the 768-dimensional EmoSpace embedding and mapped through $896\!\rightarrow\!1024\!\rightarrow\!768$.  Neither module uses layer normalization or dropout.

\noindent\textbf{Perceptual bins.}
The residual adapter receives continuous HSV means rather than discrete bin indices.  For dataset summaries and optional verbal low/mid/high descriptions, we partition the $[0,1]$-scaled training-set HSV statistics at their one-third and two-third quantiles.  The brightness thresholds are 0.4248 and 0.5636, and the saturation thresholds are 0.2563 and 0.4053.  Values at or below the lower threshold are assigned to the low bin, values at or above the upper threshold to the high bin, and the remaining values to the middle bin.

\noindent\textbf{Optimization details.}
We optimize only the controller being trained with AdamW ($\beta_1=0.9$, $\beta_2=0.999$, weight decay $10^{-2}$, and $\epsilon=10^{-8}$), a constant learning rate of $10^{-3}$ without warmup, and one epoch of FP16 training.  The per-GPU batch size is 1 and no gradient accumulation is used.  The categorical mapper is trained on four GPUs (global batch size 4) for 80,000 optimizer steps.  After freezing this mapper, the perceptual residual adapter is trained for 53,334 optimizer steps, with perceptual hidden dimension 128 and residual scale $\lambda=0.1$.  Both runs use seed 520; the EmoGen objective coefficients are \texttt{attr\_rate}$=0.01$ and \texttt{emo\_rate}$=0$.

\subsection{Continuous Branch}
\label{sec:supp_training_continuous}
\noindent\textbf{Locked-prefix prompt pairs.}
For continuous control, we construct prompt pairs with a shared semantic prefix.  Each pair contains a neutral prompt $P_0$ and a target prompt $P_t$ that preserves the same scene/object content while adding affective and perceptual visual cues.  A frozen SDXL text encoder produces token embeddings and pooled embeddings:
\begin{equation}
  (E_0,g_0)=T(P_0), \qquad (E_t,g_t)=T(P_t).
\end{equation}
The target residuals are therefore
\begin{equation}
  \Delta E^{*}=E_t-E_0, \qquad \Delta g^{*}=g_t-g_0.
\end{equation}
Because the semantic prefix is locked, these residuals mainly supervise affective--perceptual displacement rather than content replacement.

\noindent\textbf{Axis residual transformer.}
The continuous controller receives the neutral embedding and a five-axis control vector
\begin{equation}
  c=[V,A,D,B,S],
\end{equation}
where $V,A,D$ denote valence, arousal, and dominance, and $B,S$ denote brightness and saturation.  The residual transformer predicts per-axis token residuals and a pooled residual:
\begin{equation}
  \{\Delta E_V,\Delta E_A,\Delta E_D,\Delta E_B,\Delta E_S\},\Delta g
  = R_\theta(E_0,g_0,c).
\end{equation}
The final conditioning used by the frozen SDXL generator is
\begin{equation}
  E_{\mathrm{ctrl}}=E_0+\sum_i \Delta E_i, \qquad
  g_{\mathrm{ctrl}}=g_0+\Delta g.
\end{equation}

\noindent\textbf{Architecture and control normalization.}
The continuous controller contains 12 GPT-2-style transformer blocks with hidden size 768 and 12 attention heads.  The $2{,}048$-dimensional SDXL token embeddings are projected to the transformer hidden space, and the predicted pooled residual is projected back to the $1{,}280$-dimensional SDXL pooled-conditioning space.

For a raw VAD score $x\in[1,9]$, the normalized control is
\begin{equation}
  \widehat{x}_{\mathrm{VAD}}=\frac{3}{4}(x-5)\in[-3,3].
\end{equation}
Brightness and saturation are obtained from mean HSV value and saturation on a $[0,100]$ scale and normalized as
\begin{equation}
  \widehat{x}_{\mathrm{HSV}}
  =\operatorname{clip}\!\left(0.06(x-50),-3,3\right).
\end{equation}

\noindent\textbf{Embedding-residual objective.}
Following the objective defined in the main paper, we combine token-embedding reconstruction, pooled-embedding reconstruction, auxiliary axis regression, and zero-control consistency.  We report all coefficients after normalizing the token-reconstruction coefficient to one.  The coefficients of $(L_{\mathrm{tok}},L_{\mathrm{pool}},L_{\mathrm{axis}},L_0)$ are $(1,0.125,0.5,0.05)$; equivalently, $\lambda_p=0.125$, $\lambda_a=0.5$, and $\lambda_0=0.05$ in the main-paper continuous objective.  The auxiliary term regresses the five requested controls from the controller representation, while the zero-control term applies both token and pooled MSE and requires a zero control vector to reproduce the neutral conditioning.

Each prompt pair receives a quality score $q_i$ that combines caption validity, image--text similarity, image quality, annotation confidence, affective consistency, and prompt-rewrite confidence.  Within a minibatch $\mathcal{B}$, we normalize and clip the sample weights as
\begin{equation}
  \widetilde q_i=\min\!\left(10,
  \frac{q_i}{|\mathcal{B}|^{-1}\sum_{j\in\mathcal{B}}q_j}\right).
\end{equation}
For a per-sample mean-squared error $\ell_i$, the corresponding weighted reconstruction term is $|\mathcal{B}|^{-1}\sum_{i\in\mathcal{B}}\widetilde q_i\ell_i$.  At inference time, each axis is separately addressable at the input, allowing VA comparisons with EmotiCrafter as well as the additional D/B/S controls enabled by EmoScene.

\noindent\textbf{Optimization details.}
The reported checkpoint is initialized from a controller trained with the same locked-prefix objective.  Its final training run uses AdamW ($\beta_1=0.9$, $\beta_2=0.999$, weight decay $10^{-5}$, and $\epsilon=10^{-8}$) with a fixed learning rate of $5\times10^{-5}$ for 60 epochs (14,580 optimizer steps).  We use BF16 precision, eight GPUs, a per-GPU batch size of 64 (global batch size 512), and one optimizer update per minibatch without gradient accumulation.  No learning-rate scheduler or warmup is used; the validation split is 1\% and the seed is 0.

\subsection{Why Conditioning-Space Residuals?}
The two branches use different generators and conditioning formats, but both follow the same principle: learn residuals in the representation consumed by a frozen diffusion model.  This design has three practical advantages.  First, it avoids retraining large generators.  Second, it separates affect control from image-quality optimization, making the evaluation of controllability cleaner.  Third, it provides a common language for categorical affect, continuous VAD, and perceptual attributes: all of them are implemented as controllable displacements in conditioning space.

\section{Experimental Setup and Extended Results}
\label{sec:supp_experiments}

\subsection{Implementation and Generation Protocols}
\label{sec:supp_generation_protocols}
\noindent\textbf{Categorical generation.}
We follow the EmoGen-style EICG protocol: images are generated from emotion categories without input-image conditioning.  We use eight emotion categories, 50 generated images per category, resolution $512\times512$, 50 denoising steps, guidance scale 7.5, and matched seed schedules across methods.  The prompting baselines use either a fixed emotion-verbalization template with SDXL/FLUX.1 or GPT-5.5 to rewrite the same emotion-only request before generation with the corresponding frozen backbone.  The learned systems are the original EmoGen mapper trained on EmoSet, our paper-level CoEmoGen-style reproduction, the EmoGen mapper trained on EmoScene, and our full categorical controller with the perceptual residual adapter.

\noindent\textbf{Continuous generation.}
For VA control, we evaluate on 132 neutral prompts and a full $5\times5$ VA grid, resulting in 3,300 images for each compared method.  This is the shared interface supported by both EmotiCrafter and AffectCtrl.  We also test whether prompting can approximate numeric control: direct SDXL/FLUX.1 baselines verbalize each requested level, while GPT-5.5 baselines rewrite the corresponding numeric-control request before generation with the same frozen backbone.  These prompting baselines use the same 132 prompts and target levels for VAD, brightness, and saturation.  AffectCtrl is evaluated separately on dominance, brightness, and saturation using five levels per axis, yielding 660 images per axis.  For AttriCtrl (FLUX.1), we use the released checkpoint and evaluate its brightness interface under the same prompt set and target levels.

\noindent\textbf{Frozen generators.}
All reported AffectCtrl experiments keep the diffusion generators frozen.  The categorical branch uses the EmoGen-style Stable Diffusion generation interface, while the continuous branch uses frozen SDXL text encoders and generator.  This keeps the comparison focused on conditioning controllability rather than generator retraining.

\subsection{Training Data and Evaluation Sources}
\label{sec:supp_data_sources}

\noindent\textbf{Controller training data.}
The categorical controller is trained once on a combined, emotion-balanced
set of 320,000 scene- and object-conditioned entries. Because data balancing
resamples some images and a source image may contribute multiple
object-conditioned entries, the training set corresponds to 152,381 unique
source images. No separate categorical validation set is used. The continuous
controller uses a deterministic 99/1 entry-level split of its prompt-pair
dataset. The resulting training and validation sets contain 124,805 and 1,262
prompt pairs, corresponding to 77,393 and 1,259 unique source images,
respectively.

\noindent\textbf{Evaluator training data.}
The emotion evaluator is trained on 118,954 images, with 7,291 images used for
validation and 19,066 images used for internal testing. The VAD evaluator is
trained on 23,346 images, with scene-disjoint validation and test sets
containing 3,019 and 2,434 images, respectively.

\noindent\textbf{Evaluation prompt source.}
The continuous-control evaluation uses 132 unique neutral prompts selected
from the EmotiCrafter prompt pool. The set comprises 66 FindingEmo captions,
60 captions from the EMOTIC training set, and 6 OASIS captions. Only the
caption text is used as input to the generation pipeline.

\subsection{Metric Definitions}
\label{sec:supp_metrics}
\noindent\textbf{Categorical metrics.}
We report emotion accuracy (Emo-A), target-emotion confidence, semantic confidence (Sem-C), and affective control score (ACS).  Emo-A and confidence are computed using an independent CLIP-based emotion evaluator trained on held-out EmoSet/EmoScene data.  For each generated image, Sem-C is the larger of the maximum posterior probabilities produced by fixed Places365 scene and ImageNet object classifiers; we average this value over images.  ACS is the mean of Emo-A/100, target-emotion confidence, and Sem-C.

\noindent\textbf{Continuous metrics.}
For continuous control, we use independent VAD and HSV evaluators shared across all generated images.  Control errors are reported on the normalized $[-3,3]$ scale.  When evaluator predictions are produced on the original $[1,9]$ scale, we convert errors by multiplying by $3/4$, because $x_{1:9}=5+\frac{4}{3}x_{[-3,3]}$.  We additionally report Pearson correlation between the requested control level and the evaluator prediction, which measures the consistency and strength of the requested control direction.

\noindent\textbf{Additional main-table observations.}
In the categorical comparison, direct prompting reaches 71.00\% Emo-A with SDXL and 73.25\% with FLUX.1.  GPT-5.5 rewriting mainly improves semantic consistency, raising ACS to 0.684 and 0.697, respectively.  The CoEmoGen-style reproduction reaches 79.25\% Emo-A, 0.669 Sem-C, and 0.728 ACS; its semantic confidence exceeds EmoGen's 0.603, but its categorical accuracy remains lower.  For continuous control, prompting follows explicit brightness and saturation cues more readily than dominance, and GPT-5.5 improves arousal control for both backbones.  The interface-specific roles of the baselines are therefore complementary: EmotiCrafter measures shared VA control, AttriCtrl measures released brightness-only control, and the prompting rows test whether modern generators can follow verbalized numeric requests without a learned controller.

\begin{table}[!t]
\centering
\small
\resizebox{\columnwidth}{!}{%
\begin{tabular}{lccccc}
\toprule
Method & V Corr.$\uparrow$ & A Corr.$\uparrow$ & D Corr.$\uparrow$ & B Corr.$\uparrow$ & S Corr.$\uparrow$ \\
\midrule
EmotiCrafter (SDXL)~\cite{dang2025emoticrafter} & 0.701 & 0.138 & -- & -- & -- \\
AttriCtrl (FLUX.1)~\cite{chen2025attrictrl} & -- & -- & -- & 0.657 & -- \\
Ours & \textbf{0.765} & \textbf{0.673} & \textbf{0.756} & \textbf{0.762} & \textbf{0.718} \\
\bottomrule
\end{tabular}}
\caption{Supplementary Pearson correlation analysis for continuous control.  EmotiCrafter exposes VA only; AttriCtrl (FLUX.1) exposes brightness control only.}
\label{tab:supp_corr}
\end{table}

Prompting exhibits axis-specific behavior: FLUX.1 direct prompting reaches a valence correlation of 0.601, while SDXL direct prompting reaches 0.836 on saturation.  GPT-5.5 rewriting raises arousal correlation to 0.404 with SDXL and 0.388 with FLUX.1, but dominance correlations remain only 0.201 and 0.233, respectively.  For AttriCtrl (FLUX.1), monotonic ordering is strong (Mono=0.970) and brightness correlation is positive (Corr=0.657), but its outputs occupy a compressed subrange of the 1--9 brightness scale, increasing MAE despite mostly correct ordering across target levels.

\subsection{Cross-Axis Response Analysis}
\label{sec:supp_cross_axis}
The target correlations above do not reveal how the other measured dimensions respond when one requested control changes.  We therefore evaluate every generated image along all five dimensions and compute
\begin{equation}
  M_{jk}=\operatorname{corr}(c_j,\widehat y_k),
\end{equation}
where rows $j\in\{V,A,D,B,S\}$ denote the manipulated input and columns $k\in\{V,A,D,B,S\}$ denote the measured VAD/HSV response.  Each row uses the same 132 prompts and five nominal levels, giving 660 observations per manipulated axis.  The V and A rows are the $A=0$ and $V=0$ slices of the existing joint VA grid, respectively; D/B/S use their isolated sweeps.  Thus, unlike Table~\ref{tab:supp_corr}, which reports V/A correlations over the complete joint grid, this analysis isolates one requested input at a time.  The remaining requested inputs are fixed at their neutral values, which does not assume that their measured outputs remain unchanged.

\begin{figure*}[!t]
  \centering
  \includegraphics[width=0.92\textwidth]{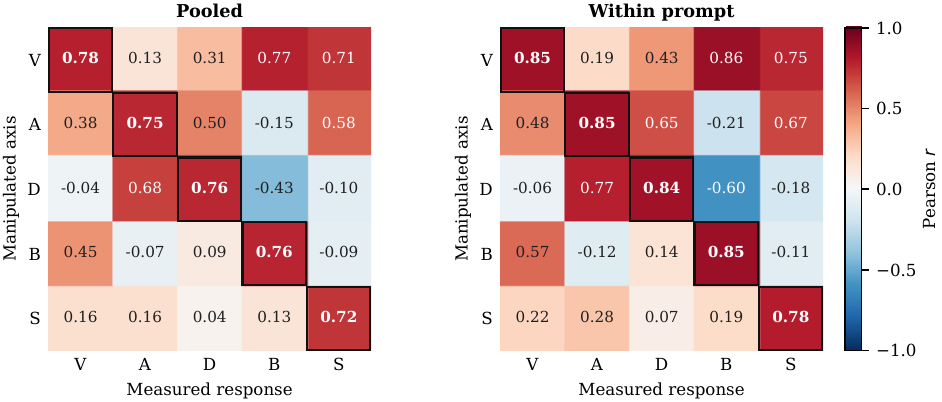}
  \caption{\textbf{Five-axis cross-axis response analysis.} Rows denote the manipulated requested control and columns denote the measured VAD/HSV response; black outlines mark target-axis entries. The pooled matrix aggregates all prompts directly. The within-prompt matrix removes each prompt's mean response across its five control levels, isolating fixed-content trajectories. Strong diagonal responses coexist with structured affect--perception coupling, so the result supports separately addressable target directions rather than complete disentanglement.}
  \label{fig:supp_cross_axis}
\end{figure*}

Figure~\ref{fig:supp_cross_axis} reports pooled correlations and a within-prompt analysis that centers each response over the five levels of the same prompt before computing $r$.  Prompt-level bootstrap with 5,000 replicates preserves the five repeated levels and gives 95\% confidence intervals.  In the pooled analysis, every row has its largest response on the requested target, with mean diagonal correlation $0.753$ [0.739, 0.769], compared with mean absolute off-axis response $0.299$ [0.287, 0.312].  The within-prompt values are $0.835$ [0.826, 0.845] and $0.377$ [0.361, 0.394], respectively.  The controls are therefore responsive but not fully orthogonal: valence strongly co-varies with brightness ($r=0.772$) and saturation ($r=0.714$), while dominance co-varies with arousal ($r=0.683$) and inversely with brightness ($r=-0.433$).  These structured responses are compatible with the affect--perception associations analyzed in Sec.~S3, but do not establish that the controller has recovered independent causal visual factors.

\subsection{Continuous-Objective Ablation}
We evaluate the contribution of each continuous-objective term through a matched retraining suite.  Full, w/o $L_{\mathrm{pool}}$, w/o $L_{\mathrm{axis}}$, and w/o $L_0$ use the same training data, initialization seed, and optimization schedule.  Evaluation fixes the same 132 prompts, requested control levels, generation seeds, sampling configuration, inference calibration, and VAD/HSV evaluators across all variants.  Each variant produces 3,300 images for the full $5\times5$ VA grid, 660 for dominance, and 1,320 for brightness and saturation.  The Full row in Table~\ref{tab:supp_continuous_ablation} is the reference trained alongside the ablations; it is a matched retraining rather than the primary checkpoint reported in the main comparison table.

\begin{table*}[!t]
\centering
\small
\setlength{\tabcolsep}{7pt}
\begin{tabular}{lcccccc}
\toprule
\multicolumn{7}{c}{\textbf{Panel A: Pearson Correlation ($\uparrow$)}} \\
\midrule
Variant & V & A & D & B & S & Avg. \\
\midrule
Full (matched) & 0.791 & \textbf{0.660} & \textbf{0.780} & 0.773 & \textbf{0.784} & \textbf{0.757} \\
w/o $L_{\mathrm{pool}}$ & 0.660 & 0.396 & 0.772 & 0.729 & 0.690 & 0.649 \\
w/o $L_{\mathrm{axis}}$ & 0.733 & 0.411 & 0.634 & 0.785 & 0.744 & 0.661 \\
w/o $L_0$ & \textbf{0.818} & 0.550 & 0.764 & \textbf{0.823} & 0.677 & 0.726 \\
\midrule
\multicolumn{7}{c}{\textbf{Panel B: Normalized MAE ($\downarrow$)}} \\
\midrule
Variant & V & A & D & B & S & Avg. \\
\midrule
Full (matched) & 1.095 & \textbf{1.351} & \textbf{1.104} & 1.288 & \textbf{1.268} & \textbf{1.221} \\
w/o $L_{\mathrm{pool}}$ & 1.314 & 1.614 & 1.153 & 1.332 & 1.500 & 1.383 \\
w/o $L_{\mathrm{axis}}$ & 1.166 & 1.616 & 1.341 & 1.257 & 1.317 & 1.339 \\
w/o $L_0$ & \textbf{1.029} & 1.459 & 1.177 & \textbf{1.192} & 1.504 & 1.272 \\
\bottomrule
\end{tabular}
\caption{Matched ablation of the continuous objective.  Panel A reports Pearson correlation between requested levels and evaluator predictions; Panel B reports MAE on the normalized $[-3,3]$ scale.  Averages are computed over V/A/D/B/S.  Bold denotes the best value in each column.}
\label{tab:supp_continuous_ablation}
\end{table*}

The complete objective gives the strongest aggregate result, with average correlation 0.757 and average MAE 1.221.  Removing pooled reconstruction produces the largest aggregate degradation, reducing average correlation by 0.108 and increasing average MAE by 0.161.  Removing axis regression most clearly affects arousal and dominance, whose correlations fall from 0.660 and 0.780 to 0.411 and 0.634.  Removing zero-control consistency improves isolated V/B scores but lowers arousal and saturation correlations, yielding a worse five-axis average.  The results therefore support the complete objective as a balanced multi-axis design rather than implying that every term improves every axis independently.

\subsection{Per-Emotion Categorical Results}
Table~\ref{tab:supp_per_emotion_categorical} expands the learned-method rows of the main categorical comparison into per-emotion results.  For CoEmoGen~\cite{yuan2026coemogen}, the official authors did not release runnable training code or pretrained weights at the time of our experiments.  We therefore implemented a paper-level CoEmoGen-style reproduction rather than presenting an official CoEmoGen result.  We train the reproduction on the EmoScene training split using sentence-level emotion-focused captions, a one-hot emotion mapper, CLIP visual-perception fusion, two polarity-shared LoRAs, and eight emotion-specific LoRAs on Stable Diffusion v1.5.

We train the reproduction for 24,000 optimizer steps with BF16 precision, AdamW ($\beta_1=0.9$, $\beta_2=0.999$), learning rate $10^{-3}$, zero weight decay, HiLoRA rank 4 and scale 4, per-device batch size 1, no gradient accumulation, and seed 0.  For evaluation, we generate 50 images per emotion at $512\times512$ resolution using 50 denoising steps, guidance scale 7.5, and the same per-emotion seed schedule as the other learned categorical methods.  We score all 400 images with the common eight-way emotion evaluator and the same Places365/ImageNet semantic-confidence evaluator used in the main categorical comparison.  Table~\ref{tab:supp_per_emotion_categorical} reports the complete per-class comparison.

\begin{table*}[!t]
\centering
\small
\setlength{\tabcolsep}{5pt}
\begin{tabular}{lcccccccc}
\toprule
\multicolumn{9}{c}{\textbf{Panel A: Emotion Accuracy and Target-Emotion Confidence}} \\
\midrule
& \multicolumn{2}{c}{EmoGen} & \multicolumn{2}{c}{CoEmoGen-style$^{\dagger}$} & \multicolumn{2}{c}{EmoGen + EmoScene} & \multicolumn{2}{c}{Ours} \\
\cmidrule(lr){2-3}\cmidrule(lr){4-5}\cmidrule(lr){6-7}\cmidrule(lr){8-9}
Emotion & Emo-A$\uparrow$ & Conf.$\uparrow$ & Emo-A$\uparrow$ & Conf.$\uparrow$ & Emo-A$\uparrow$ & Conf.$\uparrow$ & Emo-A$\uparrow$ & Conf.$\uparrow$ \\
\midrule
Amusement   & 96.00 & 0.879 & 96.00 & 0.835 & \textbf{100.00} & \textbf{0.924} & 90.00 & 0.850 \\
Awe         & 90.00 & 0.810 & \textbf{98.00} & \textbf{0.956} & 92.00 & 0.818 & \textbf{98.00} & 0.867 \\
Contentment & 78.00 & 0.588 & 90.00 & 0.765 & 86.00 & 0.757 & \textbf{96.00} & \textbf{0.862} \\
Excitement  & \textbf{92.00} & 0.816 & 90.00 & \textbf{0.845} & 86.00 & 0.725 & 82.00 & 0.798 \\
Anger       & \textbf{42.00} & \textbf{0.350} & 18.00 & 0.202 & 20.00 & 0.302 & 38.00 & 0.345 \\
Disgust     & 86.00 & 0.775 & 86.00 & 0.810 & 96.00 & 0.889 & \textbf{98.00} & \textbf{0.927} \\
Fear        & 74.00 & 0.627 & \textbf{96.00} & \textbf{0.819} & 76.00 & 0.617 & 92.00 & 0.783 \\
Sadness     & 84.00 & 0.788 & 60.00 & 0.547 & \textbf{96.00} & \textbf{0.889} & 92.00 & 0.883 \\
\midrule
Overall     & 80.25 & 0.704 & 79.25 & 0.722 & 81.50 & 0.740 & \textbf{85.75} & \textbf{0.789} \\
\midrule
\multicolumn{9}{c}{\textbf{Panel B: Semantic Confidence and Affective Control Score}} \\
\midrule
& \multicolumn{2}{c}{EmoGen} & \multicolumn{2}{c}{CoEmoGen-style$^{\dagger}$} & \multicolumn{2}{c}{EmoGen + EmoScene} & \multicolumn{2}{c}{Ours} \\
\cmidrule(lr){2-3}\cmidrule(lr){4-5}\cmidrule(lr){6-7}\cmidrule(lr){8-9}
Emotion & Sem-C$\uparrow$ & ACS$\uparrow$ & Sem-C$\uparrow$ & ACS$\uparrow$ & Sem-C$\uparrow$ & ACS$\uparrow$ & Sem-C$\uparrow$ & ACS$\uparrow$ \\
\midrule
Amusement   & 0.612 & 0.817 & \textbf{0.759} & 0.852 & 0.695 & \textbf{0.873} & 0.721 & 0.824 \\
Awe         & \textbf{0.722} & 0.811 & 0.582 & \textbf{0.839} & 0.507 & 0.748 & 0.633 & 0.827 \\
Contentment & 0.542 & 0.637 & \textbf{0.601} & 0.755 & 0.581 & 0.733 & 0.466 & \textbf{0.763} \\
Excitement  & 0.581 & 0.772 & 0.582 & 0.776 & 0.714 & 0.767 & \textbf{0.734} & \textbf{0.784} \\
Anger       & 0.560 & 0.443 & 0.661 & 0.348 & 0.945 & 0.482 & \textbf{0.947} & \textbf{0.557} \\
Disgust     & 0.696 & 0.777 & \textbf{0.857} & 0.842 & 0.821 & 0.890 & \textbf{0.857} & \textbf{0.922} \\
Fear        & 0.467 & 0.611 & 0.702 & \textbf{0.827} & 0.671 & 0.683 & \textbf{0.748} & 0.817 \\
Sadness     & 0.646 & 0.758 & 0.610 & 0.586 & 0.876 & \textbf{0.908} & \textbf{0.892} & 0.898 \\
\midrule
Overall     & 0.603 & 0.703 & 0.669 & 0.728 & 0.726 & 0.760 & \textbf{0.750} & \textbf{0.799} \\
\bottomrule
\end{tabular}
\caption{Per-emotion categorical comparison of learned methods under the common protocol of 50 images per emotion.  Panel A reports eight-class emotion accuracy (Emo-A, in percentage) and target-emotion confidence (Conf.); Panel B reports semantic confidence (Sem-C) and ACS, the mean of Emo-A/100, Conf., and Sem-C.  Bold indicates the best displayed value for each emotion and metric.  $^{\dagger}$Our paper-level CoEmoGen-style reproduction, not an official result; the official authors did not release runnable training code or pretrained weights at the time of our experiments.}
\label{tab:supp_per_emotion_categorical}
\end{table*}

Across the four learned methods, AffectCtrl obtains the best overall value for every metric, including 85.75\% Emo-A and 0.799 ACS.  The per-emotion breakdown also shows complementary strengths rather than uniform dominance: EmoGen is strongest on excitement and anger, the EmoScene mapper is strongest on amusement and sadness, and the CoEmoGen-style reproduction is strongest on fear while tying AffectCtrl on awe accuracy.  The reproduction improves overall semantic confidence over EmoGen (0.669 versus 0.603), consistent with sentence-level semantic guidance, but has lower overall emotion accuracy (79.25\% versus 80.25\%).  Reporting all methods under the same evaluators keeps the category-level comparison symmetric while preserving the distinction between official and locally reproduced results.

\subsection{EmoCtrl under Its Native Interface}
EmoCtrl~\cite{yang2025emoctrl} differs from the emotion-only interface used in Table~\ref{tab:supp_per_emotion_categorical}: it requires both a content condition and a target emotion.  We therefore report it separately rather than inserting it into the matched emotion-only comparison.  We use the released EmoCtrl implementation, but train the Stage-1, Stage-2, and Stage-3 weights locally because the release does not provide pretrained weights or an emotion-evaluator checkpoint.  Stage 3 uses the epoch-10 checkpoint specified by the method.

To remove content--emotion confounding, we construct a fixed content set from EmoCtrl's released test pool.  We deterministically select 25 emotion-neutral elements and 25 emotion-neutral captions, and reuse exactly the same 50 content conditions for each of the eight target emotions, yielding 400 images.  We generate at $512\times512$ resolution with 50 denoising steps, guidance scale 7.5, the method's native emotion scaling of 0.3, and the same per-emotion seed schedule used in our categorical experiments.  All images are scored by the common emotion and semantic-confidence evaluators defined above.  Table~\ref{tab:supp_emoctrl_native} reports this local reproduction under EmoCtrl's native content-conditioned interface.

\begin{table}[!t]
\centering
\small
\setlength{\tabcolsep}{4pt}
\begin{tabular}{lcccc}
\toprule
Emotion & Emo-A$\uparrow$ & Conf.$\uparrow$ & Sem-C$\uparrow$ & ACS$\uparrow$ \\
\midrule
Amusement   & 22.00 & 0.225 & 0.593 & 0.346 \\
Awe         & 52.00 & 0.430 & 0.527 & 0.492 \\
Contentment & 44.00 & 0.349 & 0.544 & 0.445 \\
Excitement  & 56.00 & 0.458 & 0.603 & 0.540 \\
Anger       & 22.00 & 0.196 & 0.605 & 0.340 \\
Disgust     & 30.00 & 0.219 & 0.560 & 0.359 \\
Fear        & 54.00 & 0.466 & 0.575 & 0.527 \\
Sadness     & 56.00 & 0.478 & 0.606 & 0.548 \\
\midrule
Overall     & 42.00 & 0.353 & 0.577 & 0.450 \\
\bottomrule
\end{tabular}
\caption{Local reproduction of EmoCtrl under its native content-conditioned interface.  The same 50 emotion-neutral content conditions are reused across all eight emotions, and all 400 images are scored with our common evaluators.  Emo-A is reported in percentage.  This result is kept separate from the emotion-only comparison because the input interfaces are not identical.}
\label{tab:supp_emoctrl_native}
\end{table}

The fixed-content protocol yields 42.00\% Emo-A and 0.450 ACS.  Performance varies across categories, reaching 56.00\% for excitement and sadness but 22.00\% for amusement and anger.  These values document the behavior of the released method under our evaluator without treating the result as an identical-interface comparison or as an official pretrained-weight result.

\subsection{Human Study Protocols}
We conduct three human studies corresponding to the three controllability claims in the main paper.  The first study compares categorical emotion generation against EmoGen.  The second study compares VA controllability against EmotiCrafter.  The third study evaluates the full dual-space interface of AffectCtrl, including VAD, brightness, and saturation, using a yes/no judgment protocol.

\noindent\textbf{Blinding, presentation order, and response handling.}
In both pairwise studies, method identities are hidden from annotators.  The categorical study independently randomizes the left--right order for each trial and provides an ``unsure'' option, while the VA study independently randomizes the top--bottom order and provides a ``tie/unable'' option.  The full dual-space study provides ``yes,'' ``no,'' and ``unable'' responses; only yes/no responses are counted as valid judgments, with ``unable'' and unanswered entries excluded.

\noindent\textbf{Categorical preference study.}
Annotators are shown paired images generated for the same target emotion and choose which image better expresses the target affect while preserving plausible scene content.  The study contains 160 image pairs, three annotators, and 478 valid votes out of 480 expected votes.  Table~\ref{tab:supp_human_cat} reports both vote-level preference and pair-level majority win rate.  The only clear failure case is excitement, where EmoGen receives stronger preference; this is consistent with the difficulty of visually separating excitement from other high-arousal positive emotions.

\begin{table*}[!t]
\centering
\small
\setlength{\tabcolsep}{5pt}
\begin{tabular}{lcccccc}
\toprule
Emotion & \# Pairs & Valid Votes & Ours Votes & EmoGen Votes & Ours Vote Pref.$\uparrow$ & Ours Pair Win$\uparrow$ \\
\midrule
Overall & 160 & 478 & 357 & 121 & 74.7\% & 82.5\% \\
Amusement & 20 & 60 & 42 & 18 & 70.0\% & 85.0\% \\
Anger & 20 & 59 & 51 & 8 & 86.4\% & 100.0\% \\
Awe & 20 & 60 & 50 & 10 & 83.3\% & 100.0\% \\
Contentment & 20 & 60 & 53 & 7 & 88.3\% & 95.0\% \\
Disgust & 20 & 60 & 52 & 8 & 86.7\% & 95.0\% \\
Excitement & 20 & 60 & 9 & 51 & 15.0\% & 5.0\% \\
Fear & 20 & 59 & 52 & 7 & 88.1\% & 100.0\% \\
Sadness & 20 & 60 & 48 & 12 & 80.0\% & 80.0\% \\
\bottomrule
\end{tabular}
\caption{Human preference study for categorical emotion generation against EmoGen.  Pair win is computed by majority vote over annotators for each image pair.}
\label{tab:supp_human_cat}
\end{table*}

\begin{figure*}[!t]
\begin{minipage}[t]{0.49\textwidth}
  \vspace{0pt}
  \centering
  \small
  \resizebox{\linewidth}{!}{%
  \begin{tabular}{lcccccc}
  \toprule
  Axis & \# Pairs & Ours & Emoti. & Tie & Vote Pref.$\uparrow$ & Pair Win$\uparrow$ \\
  \midrule
  Overall & 80 & 241 & 58 & 21 & 80.6\% & 77.5\% \\
  Valence & 40 & 124 & 30 & 6 & 80.5\% & 80.0\% \\
  Arousal & 40 & 117 & 28 & 15 & 80.7\% & 75.0\% \\
  \bottomrule
  \end{tabular}}
  \captionof{table}{Human VA controllability study against EmotiCrafter. Vote preference excludes tie votes; no-majority pairs are reported separately.}
  \label{tab:supp_human_va}
  \vspace{0.6em}
  \setlength{\tabcolsep}{3pt}
  \resizebox{\linewidth}{!}{%
  \begin{tabular}{llcccccc}
  \toprule
  Space & Axis & $N$ & Valid & Yes & No & Accept$\uparrow$ & Maj. Accept$\uparrow$ \\
  \midrule
  Overall & All & 100 & 394 & 267 & 127 & 67.8\% & 63.0\% \\
  Affective & All & 60 & 234 & 163 & 71 & 69.7\% & 65.0\% \\
  Perceptual & All & 40 & 160 & 104 & 56 & 65.0\% & 60.0\% \\
  Affective & Valence & 20 & 80 & 61 & 19 & 76.2\% & 80.0\% \\
  Affective & Arousal & 20 & 80 & 58 & 22 & 72.5\% & 70.0\% \\
  Affective & Dominance & 20 & 74 & 44 & 30 & 59.5\% & 45.0\% \\
  Perceptual & Brightness & 20 & 80 & 52 & 28 & 65.0\% & 55.0\% \\
  Perceptual & Saturation & 20 & 80 & 52 & 28 & 65.0\% & 65.0\% \\
  \bottomrule
  \end{tabular}}
  \captionof{table}{Human yes/no study for full dual-space control. Human accept is the yes rate over valid judgments; majority-accepted is the percentage of images accepted by majority vote.}
  \label{tab:supp_human_dual}
\end{minipage}
\hfill
\begin{minipage}[t]{0.49\textwidth}
  \vspace{0pt}
  \centering
  \includegraphics[width=\linewidth,trim=0 150bp 0 0,clip]{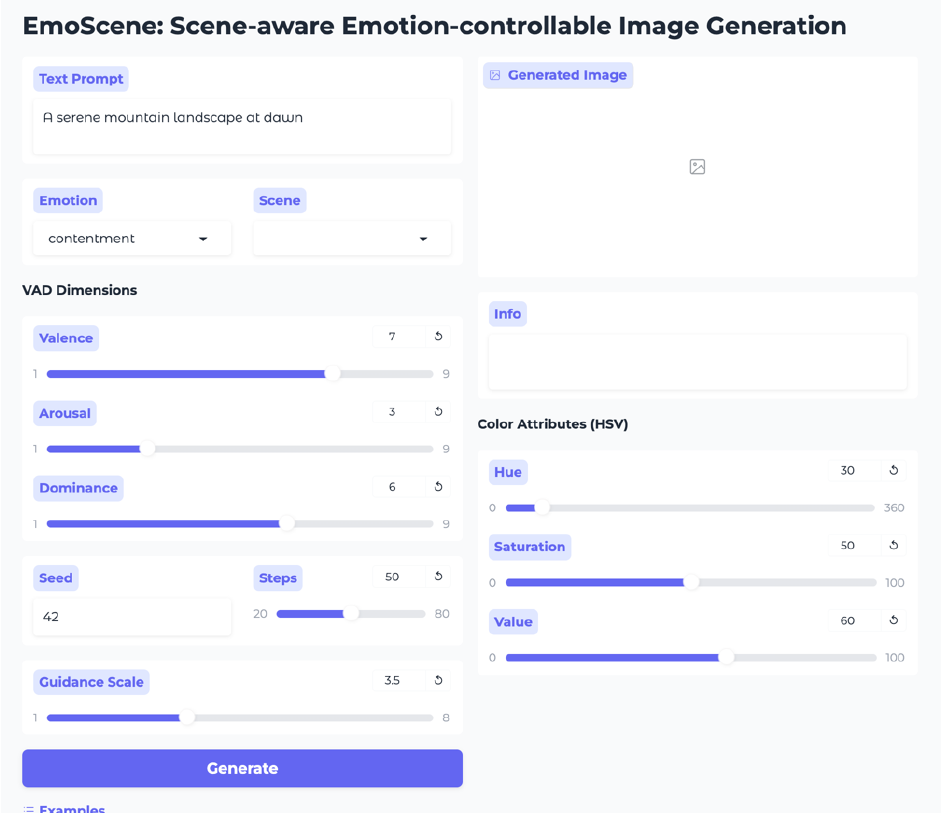}
  \captionof{figure}{\textbf{Interactive demo.}
  Dataset annotations are exposed as simple controls for categorical emotion, VAD, and perceptual adjustment, with sampling settings kept explicit for reproducibility.}
  \label{fig:gradio-ui}
\end{minipage}
\end{figure*}

\noindent\textbf{VA pairwise controllability study.}
For the EmotiCrafter comparison, annotators are shown paired images generated from the same prompt and the same target VA direction.  They choose which method better reflects the requested valence or arousal change.  Table~\ref{tab:supp_human_va} reports the overall result and the per-axis split.

\noindent\textbf{Full dual-space yes/no study.}
The third study evaluates whether images generated by AffectCtrl match a specified target control without comparing to another method.  This study covers VAD and perceptual axes.  Table~\ref{tab:supp_human_dual} shows that the full interface is generally accepted by human raters, while dominance is the hardest axis, matching our observation that dominance is visually more abstract than brightness or saturation.

\noindent\textbf{Interactive demo.}
We implement a lightweight Gradio interface for visual inspection of dataset-grounded controls.  The interface exposes a text prompt, emotion selector, VAD sliders, brightness and saturation controls, seed, denoising steps, and guidance scale.  The generated image and execution log are shown side by side, allowing users to compare different control settings under matched prompts and seeds.

\clearpage

\twocolumn[
\section{Extended Qualitative Results}
\label{sec:supp_qual}
]

\noindent\begin{minipage}{\columnwidth}
\subsection{VA Comparison with EmotiCrafter}
Figure~\ref{fig:supp-va-comparison} provides a matched qualitative comparison on the shared VA interface. Both methods use the same prompt and traverse a shared VA grid: valence increases from left to right and arousal increases from top to bottom. AffectCtrl exhibits a clearer ordered progression while preserving the lake-and-pier scene structure, complementing the automatic and human VA evaluations in the main paper.
\par\medskip
  \centering
  \includegraphics[width=\linewidth]{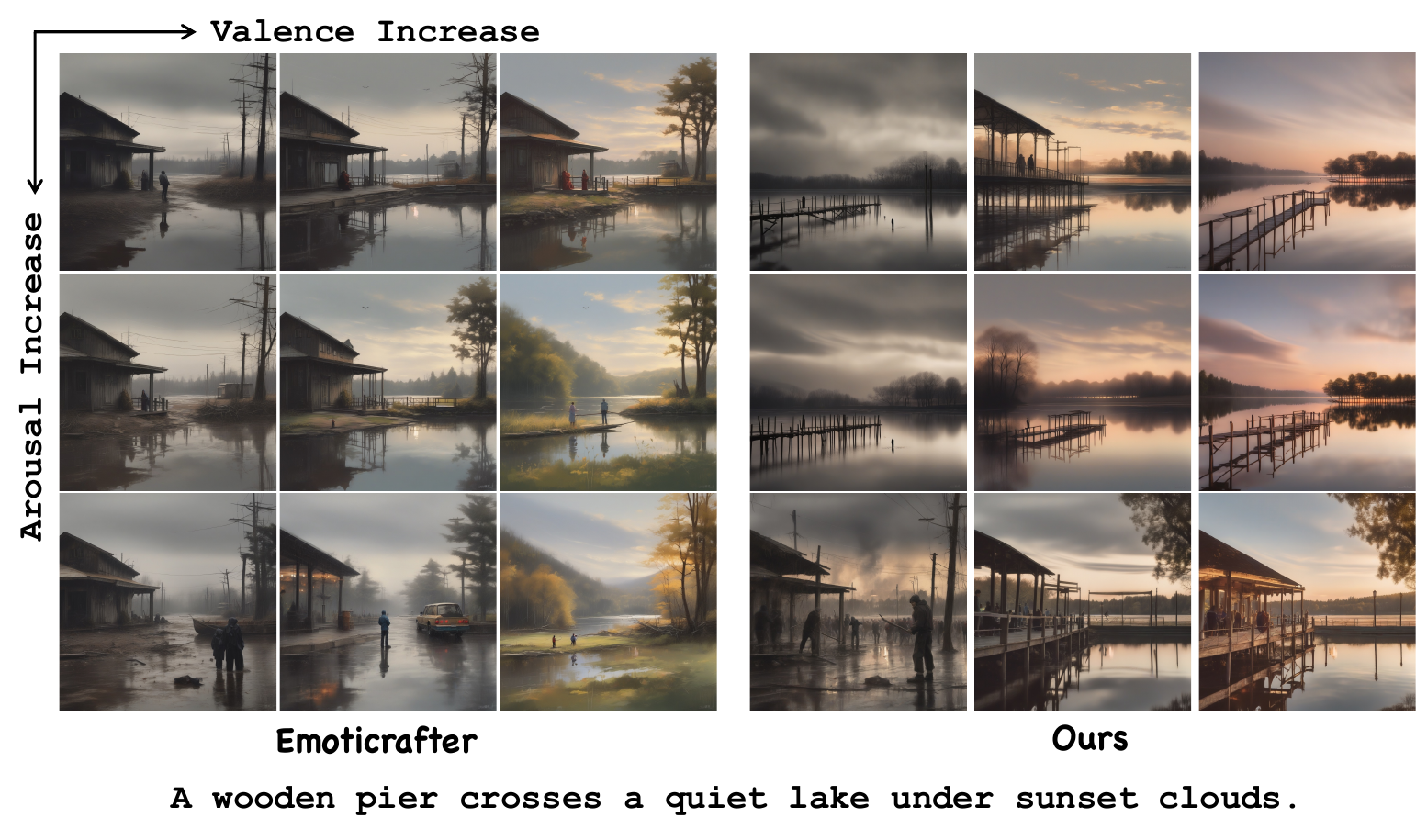}
  \captionof{figure}{\textbf{Qualitative VA comparison with EmotiCrafter.} EmotiCrafter is shown on the left and AffectCtrl on the right under a fixed prompt and matched sampling setup. Columns increase valence from left to right, and rows increase arousal from top to bottom.}
  \label{fig:supp-va-comparison}
\end{minipage}

\subsection{Categorical Emotion Generation}
\begin{figure*}[!t]
  \centering
  \includegraphics[width=0.94\textwidth]{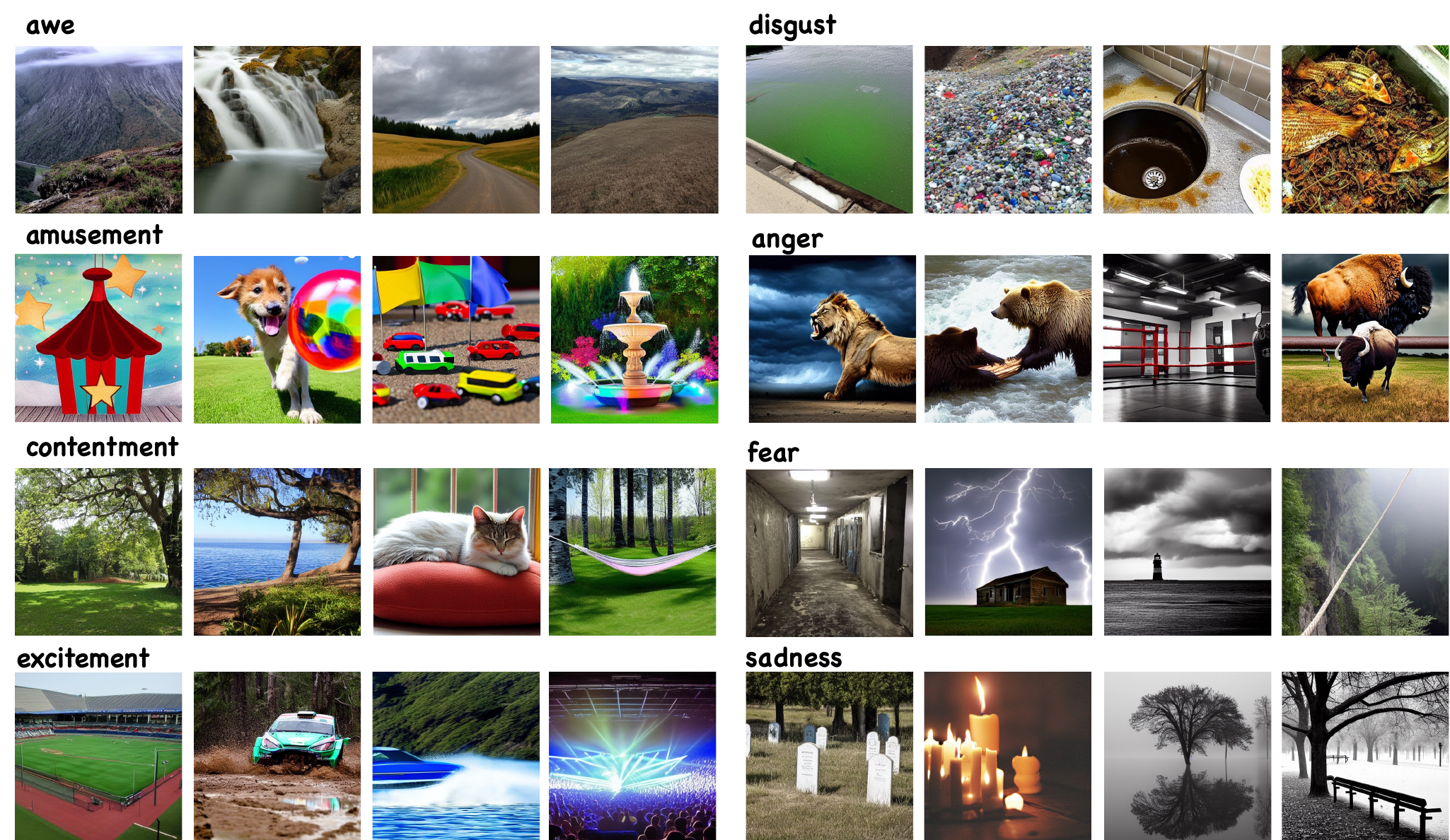}
  \caption{\textbf{Discrete emotion generation across eight target categories.}  We show four representative AffectCtrl generations for each requested emotion: awe, amusement, contentment, excitement, disgust, anger, fear, and sadness.  All images are generated by the full categorical branch, which combines an EmoScene-trained affective token with the perceptual residual adapter, using the common $512\times512$ protocol (50 denoising steps and guidance scale 7.5).  Semantic prompts and random seeds vary across samples to expose within-category scene diversity; the figure is a qualitative gallery rather than a paired baseline comparison.}
  \label{fig:supp-discrete-generation}
\end{figure*}

Figure~\ref{fig:supp-discrete-generation} complements the two-category paired comparison in the main paper by visualizing the complete eight-class output space of the AffectCtrl categorical branch.  Each category block contains four representative generations produced with the protocol in Sec.~\ref{sec:supp_experiments}.  The examples cover natural landscapes, animals, objects, built environments, and event scenes.  Across this semantic variation, the generations exhibit recognizable category-specific cues: expansive scenery for awe, playful objects and vivid colors for amusement, tranquil settings for contentment, dynamic sports and motion for excitement, contamination and decay for disgust, confrontation and force for anger, threatening environments for fear, and isolation or loss for sadness.  Because semantic prompts and random seeds vary across samples, this figure illustrates generation breadth rather than a matched-prompt baseline comparison; the controlled comparison and human preference results are reported in Table~3 of the main paper and Table~\ref{tab:supp_human_cat}, respectively.

In the matched main-paper comparison, AffectCtrl renders awe as an expansive mountain landscape rather than a tranquil settlement and renders disgust with visible litter rather than clean foliage. The methods use the same per-emotion seeds and sampling configuration, so these examples isolate target-specific visual cues; the quantitative eight-class comparison remains the primary evidence.

\subsection{Five-Axis Continuous Control Trajectories}
\begin{figure*}[!t]
  \centering
  \includegraphics[width=\textwidth]{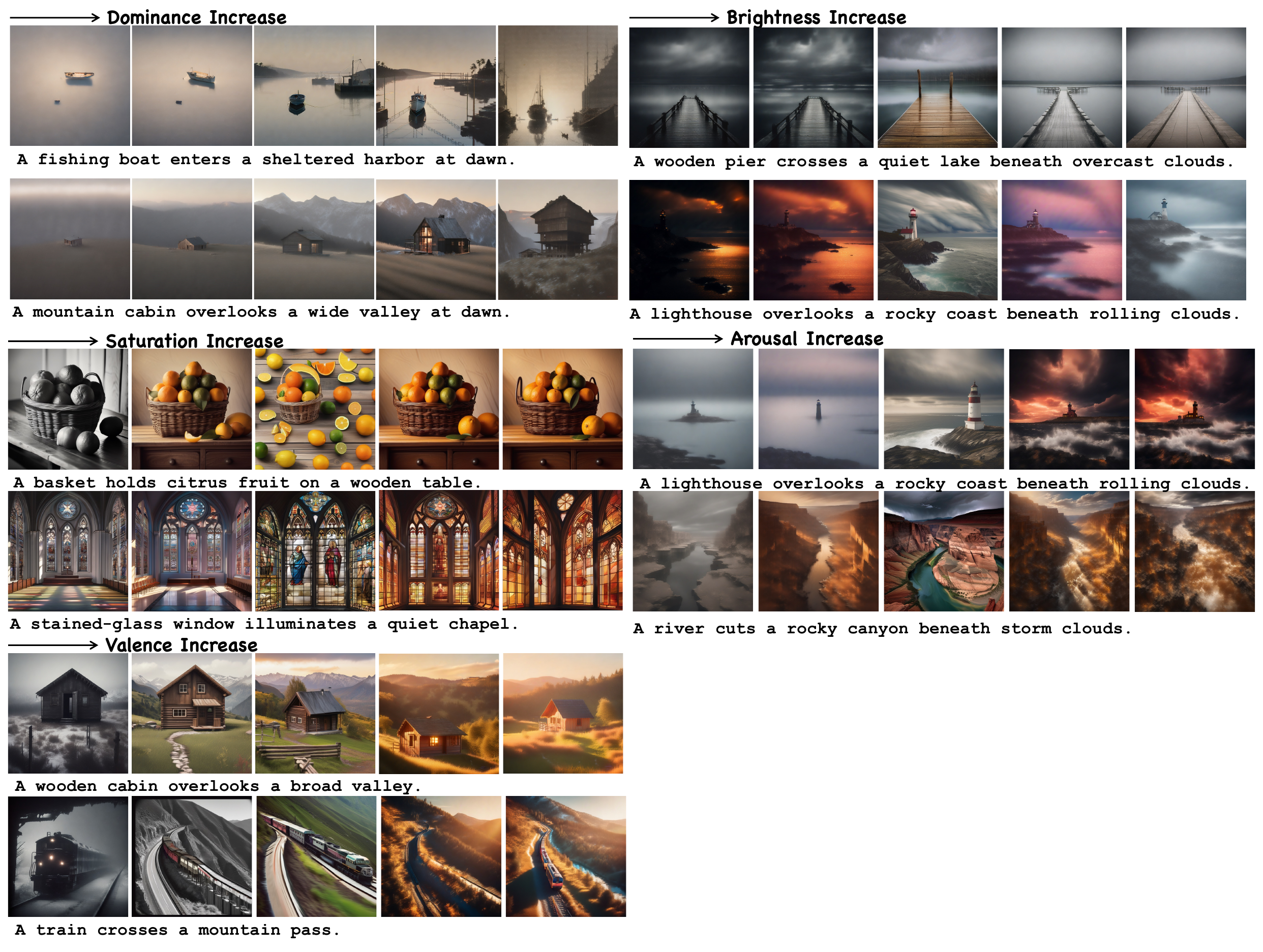}
  \caption{\textbf{Five-axis continuous dual-space control.}  Each sequence sweeps one requested axis from low to high (left to right) over the nominal normalized levels $[-3,-1.5,0,1.5,3]$, while the remaining four requested inputs are fixed at their neutral settings.  Prompts and seeds are fixed within each row.  AffectCtrl produces progressive target-axis trajectories over valence, arousal, dominance, brightness, and saturation without implying that the measured responses are fully independent.}
  \label{fig:supp-continuous-five-axis}
\end{figure*}

Figure~\ref{fig:supp-continuous-five-axis} provides a compact qualitative overview of the full AffectCtrl interface.  For each row, we hold the SVO prompt and random seed fixed, vary only the indicated control, and traverse the five nominal normalized levels $[-3,-1.5,0,1.5,3]$ from left to right.  The sequences exhibit progressive changes along the three affective dimensions (valence, arousal, and dominance) and two perceptual dimensions (brightness and saturation), while retaining recognizable scene content.  These qualitative trends complement the Pearson correlations in Table~\ref{tab:supp_corr} and the human acceptability results in Table~\ref{tab:supp_human_dual}.

\subsection{Extended Dataset Galleries}
\label{sec:supp_galleries}
\begin{figure*}[htbp]
    \centering
    \includegraphics[width=0.93\textwidth]{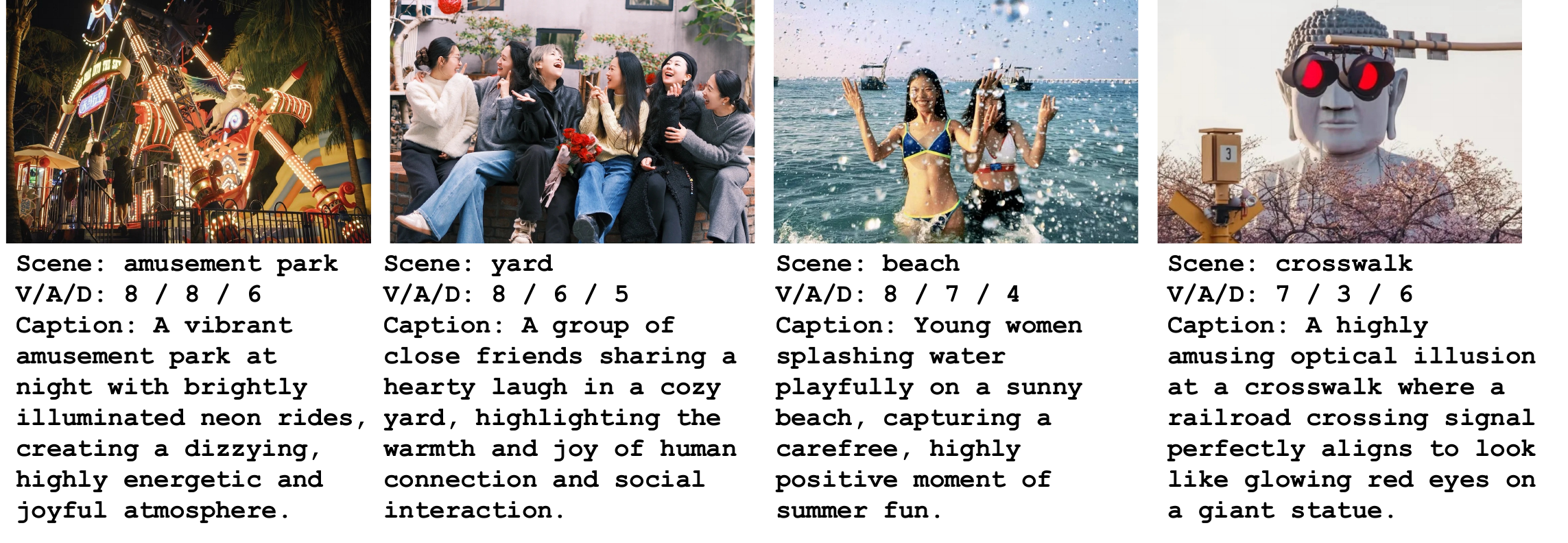}
    \vspace{0.05cm}
    \includegraphics[width=0.93\textwidth]{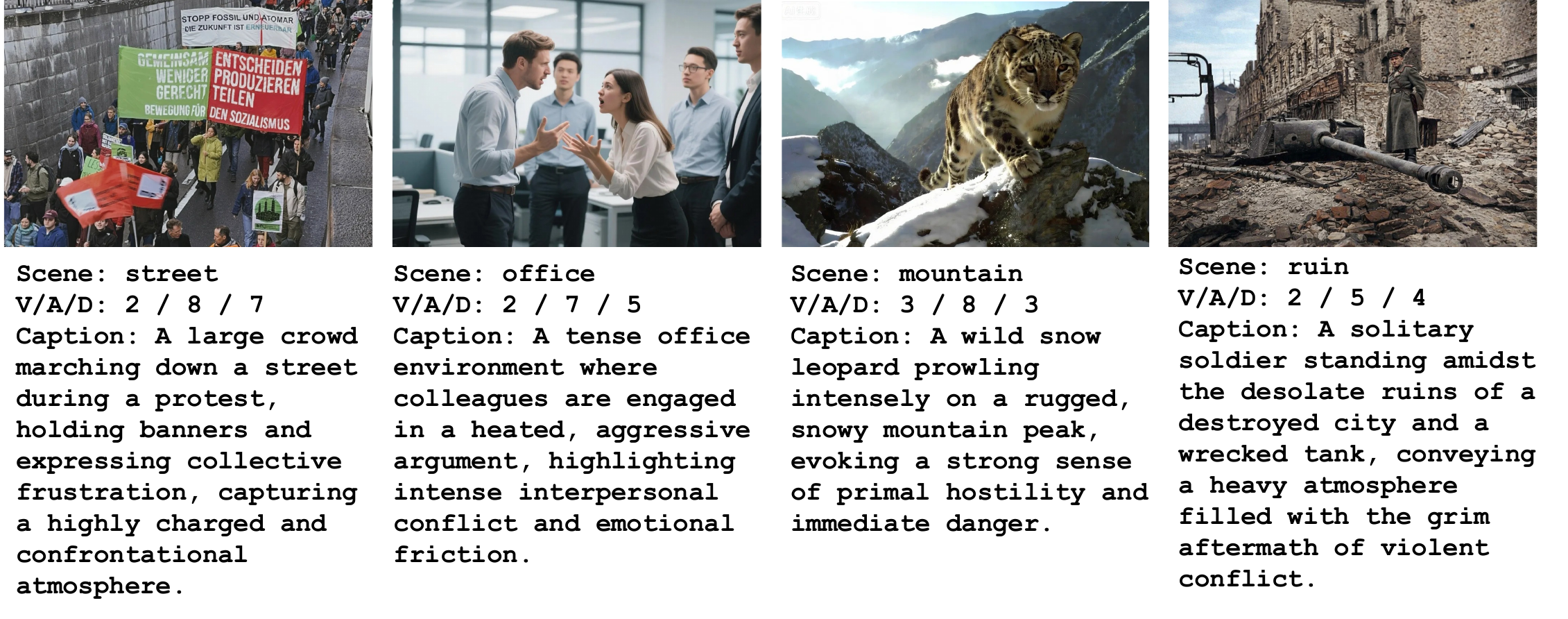}
    \vspace{0.05cm}
    \includegraphics[width=0.93\textwidth]{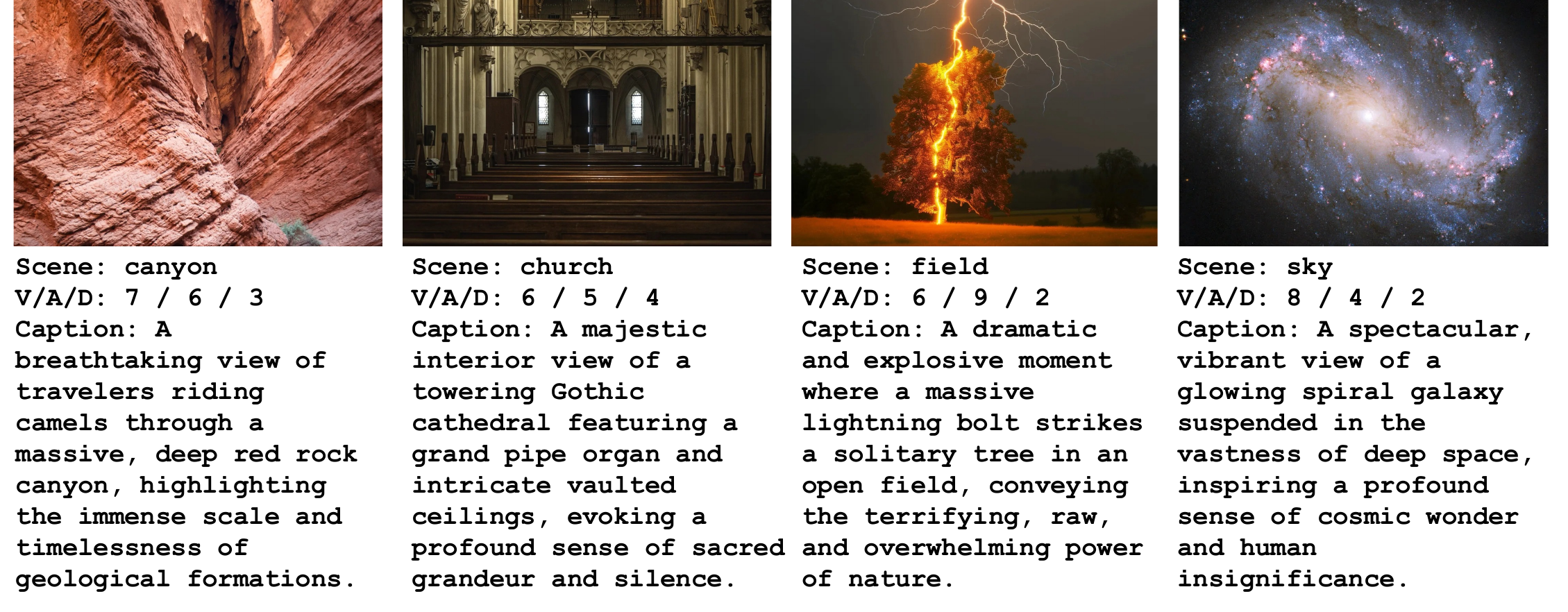}
    \caption{\textbf{Extended EmoScene Dataset Gallery (Part 1/3).} The three gallery figures cover eight target emotions plus Neutral, with VAD scores, scene labels, and captions that expose intra-class semantic and perceptual diversity. This first part shows \textbf{Amusement}, \textbf{Anger}, and \textbf{Awe} across social interactions, urban conflicts, and natural landscapes. Dominance varies with semantic context, including low dominance in many awe scenes and a wider range for anger.}
    \label{fig:gallery_1}
\end{figure*}

\begin{figure*}[htbp]
    \centering
    \includegraphics[width=0.93\textwidth]{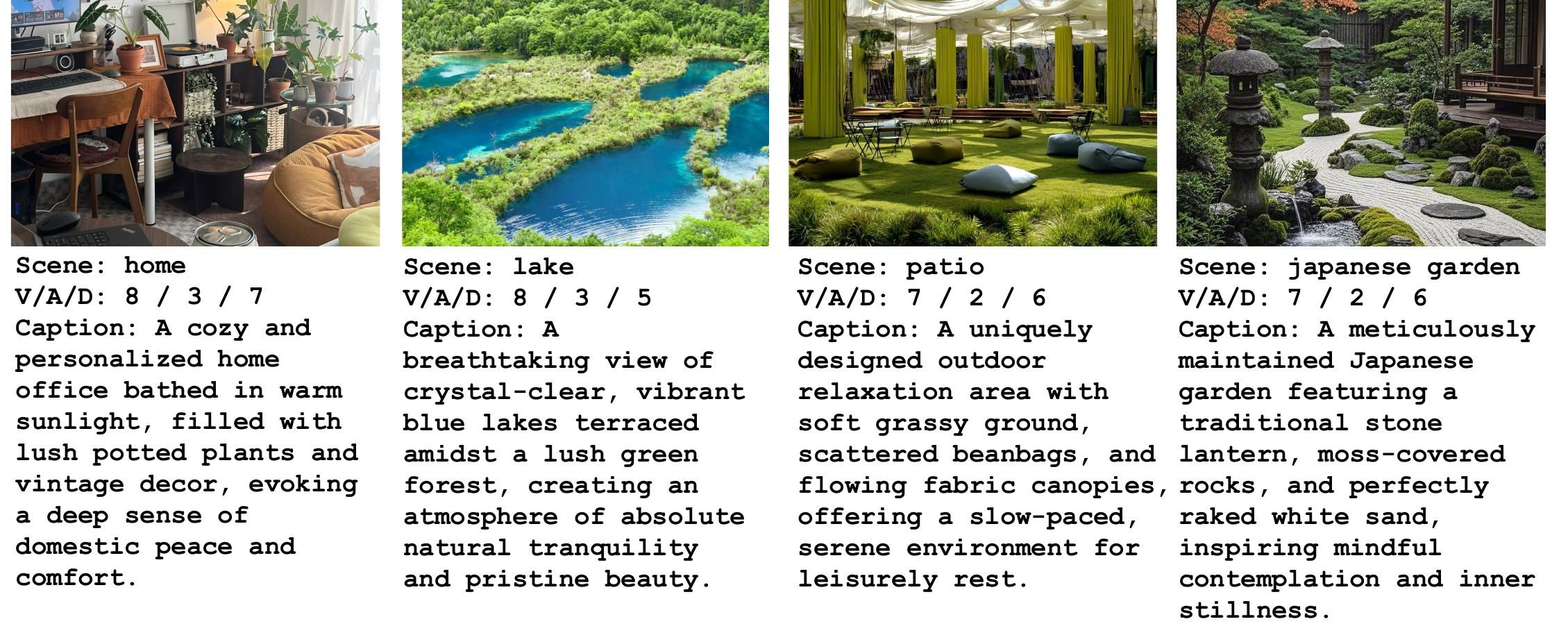}
    \vspace{0.05cm}
    \includegraphics[width=0.93\textwidth]{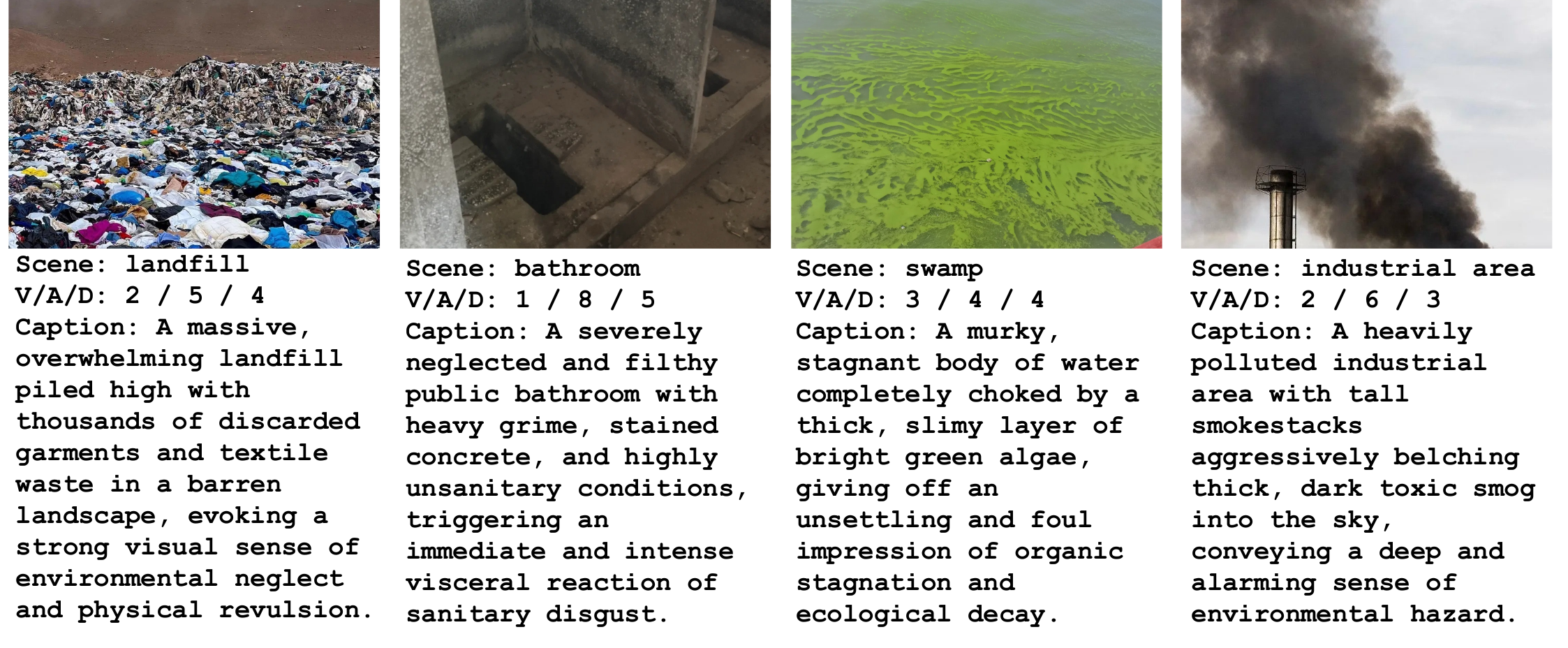}
    \vspace{0.05cm}
    \includegraphics[width=0.93\textwidth]{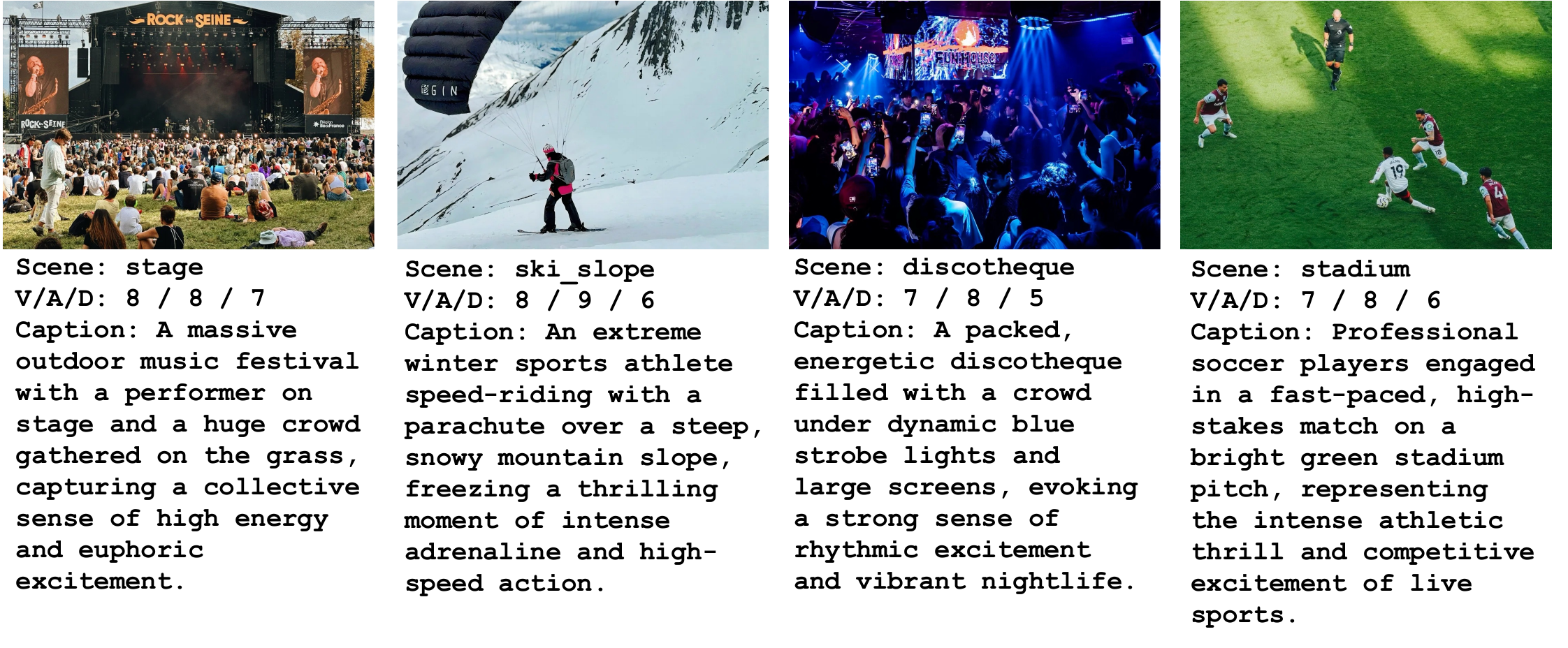}
    \caption{\textbf{EmoScene Dataset Gallery (Part 2/3).} Intra-class diversity for \textbf{Contentment}, \textbf{Disgust}, and \textbf{Excitement}. These variations highlight the dataset's ability to ground distinct activation levels and valences across entirely different scenes—from the absolute tranquility and low arousal of \textit{Contentment} to the peak dynamic energy of \textit{Excitement} and the multifaceted environmental revulsion of \textit{Disgust}.}
    \label{fig:gallery_2}
\end{figure*}

\begin{figure*}[htbp]
    \centering
    \includegraphics[width=0.93\textwidth]{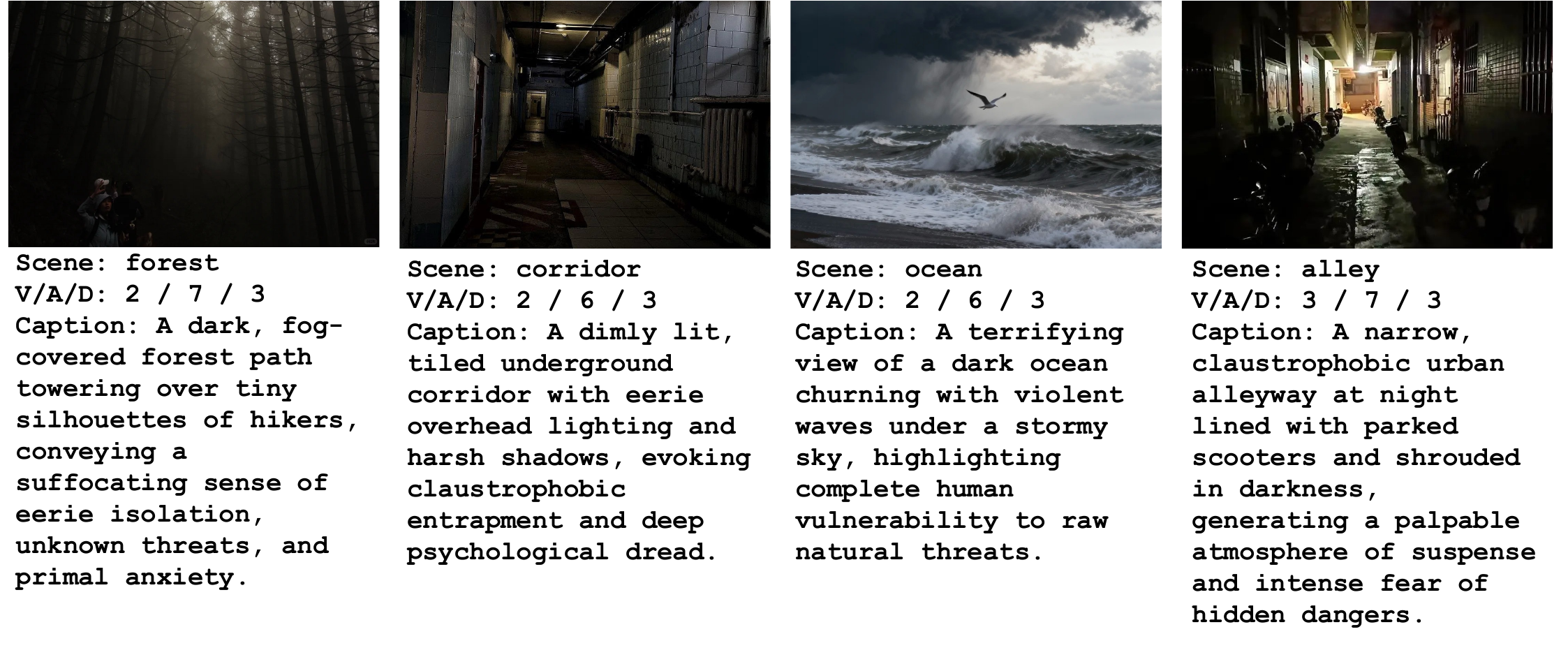}
    \vspace{0.05cm}
    \includegraphics[width=0.93\textwidth]{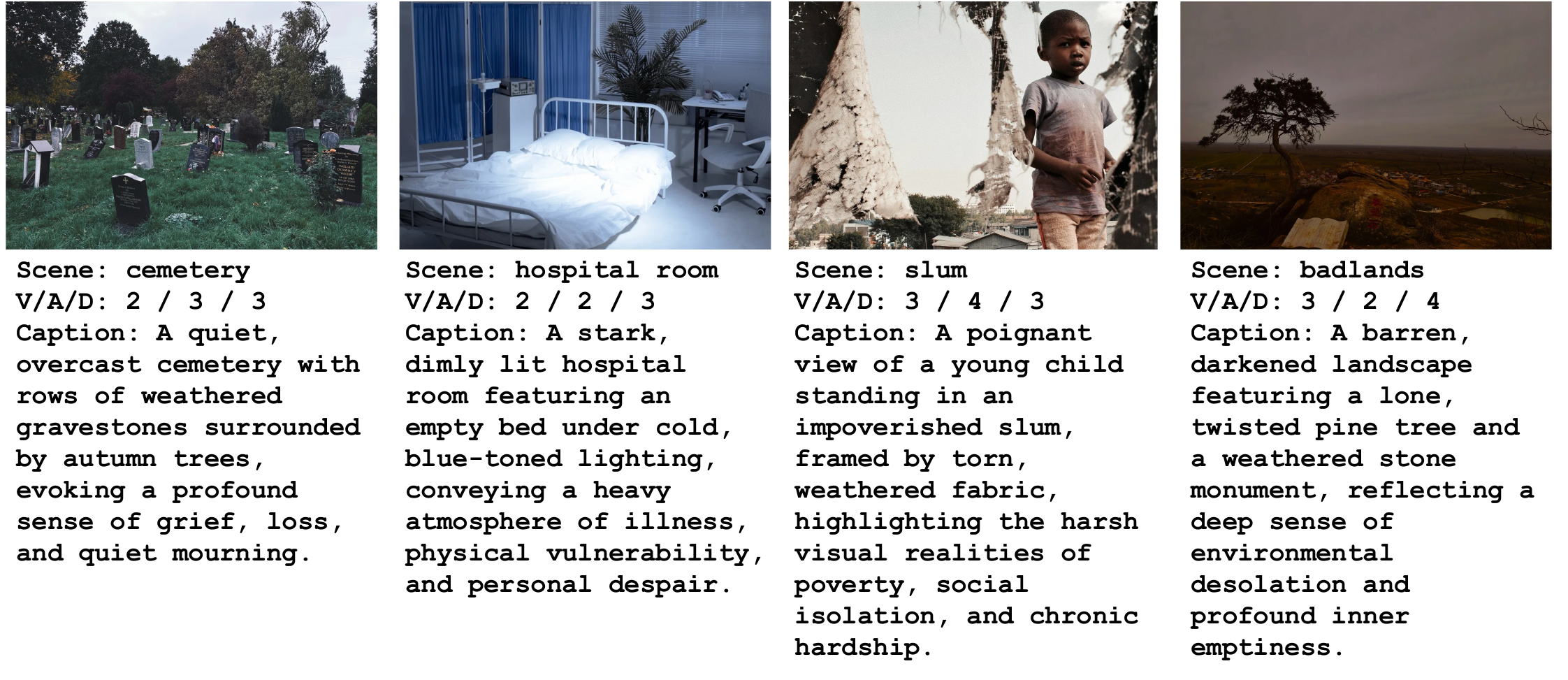}
    \vspace{0.05cm}
    \includegraphics[width=0.93\textwidth]{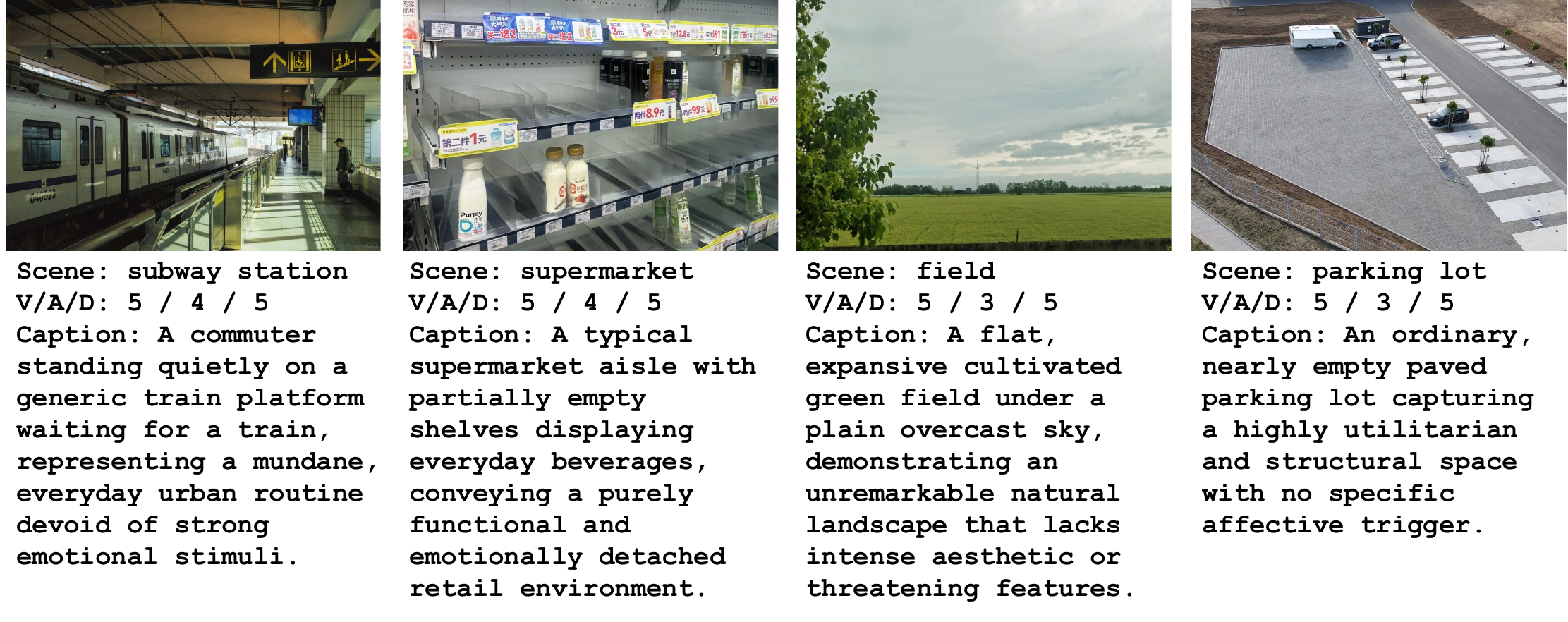}
    \caption{\textbf{EmoScene Dataset Gallery (Part 3/3).} Intra-class diversity for \textbf{Fear}, \textbf{Sadness}, and \textbf{Neutral}. While Fear and Sadness capture intensely negative affective states with varying arousal, the \textbf{Neutral} category serves as a crucial baseline. Characterized by mid-range valence, arousal, and dominance (VAD $\approx$ 5/5/5), these neutral samples ground the dataset in everyday, purely functional, and emotionally detached semantics (e.g., mundane urban routines or utilitarian spaces).}
    \label{fig:gallery_3}
\end{figure*}

Figures~\ref{fig:gallery_1}--\ref{fig:gallery_3} complement the generation results with representative EmoScene samples from eight target emotions plus Neutral. Each example includes VAD scores, a scene label, and a contextual caption, showing intra-class semantic and perceptual diversity rather than only class prototypes.

\FloatBarrier

\end{document}